%% file: main.tex
\newif\ifcomments  %
\newif\ifneurips  %
\newif\ifsupp
\commentsfalse
\neuripsfalse
\suppfalse
\documentclass{article}

\input{pkgs}

\input{macros}
\ifneurips
\usepackage{neurips_2020}
\else
\usepackage{fullpage}
\usepackage[numbers, sort, comma, square]{natbib}
\fi

\title{Privacy Amplification via Random Check-Ins}
\author{
Borja Balle\thanks{DeepMind. \texttt{bballe@google.com}} \and Peter Kairouz\thanks{Google. \texttt{\{kairouz, mcmahan, omthkkr\}@google.com}} \and H. Brendan McMahan\samethanks[2]
\and
Om Thakkar\samethanks[2]
\and
Abhradeep Thakurta\thanks{Google Research - Brain.  \texttt{\{athakurta\}@google.com}}}

\begin{document}

\maketitle
\input{abstract}
\input{intro}

\input{prelims}

\input{new_checkins}

\input{shuffling}

\input{conclusion}

\input{acks}
\bibliographystyle{abbrv}
\bibliography{reference}
\newpage
\appendix
\input{app_proofs}
\end{document}

%% file: pkgs.tex
\usepackage{amsmath,amssymb,pifont}
\usepackage[noend]{algorithmic}
\usepackage{algorithm}
\usepackage{multicol}
\usepackage{amstext}
\usepackage{amsthm}
\usepackage{multirow}
\usepackage{booktabs}
\usepackage[skip=0pt]{subcaption}
\usepackage{times}
\usepackage{lipsum}
\usepackage[shortlabels]{enumitem}
\usepackage{wrapfig}
\usepackage{array}
\usepackage{siunitx}
\usepackage{csvsimple}
\usepackage[multidot]{grffile}
\usepackage{bbm}
\usepackage{dblfloatfix}
\usepackage{hyperref}
\usepackage{makecell}
\usepackage{bbm, dsfont}
\usepackage{mathtools}
\usepackage{xcolor}
\usepackage{comment}
\usepackage[multiple]{footmisc}
\usepackage{mathrsfs}

%% file: macros.tex
\renewcommand{\epsilon}{\varepsilon}
\newcommand{\mypar}[1]{\smallskip
	\noindent{\textbf{{#1}:}}}
	
\newcommand{\SUB}[1]{\hspace{-0.15in} \textbf{#1}}
\newcommand{\mycaptionof}[2]{\captionof{#1}{#2}}

\newcommand\numberthis{\addtocounter{equation}{1}\tag{\theequation}}
\hypersetup{final}

\newcommand{\eps}{\ensuremath{\varepsilon}}

\newcommand{\calA}{\ensuremath{\mathcal{A}}}
\newcommand{\calB}{\ensuremath{\mathcal{B}}}

\newcommand{\calD}{\ensuremath{\mathcal{D}}}

\newcommand{\calS}{\ensuremath{\mathcal{S}}}

\renewcommand{\Pr}{\mathop{\mathbf{Pr}}}

\newcommand{\BiasedSampling}{\mathsf{BiasedSampling}}

\newtheorem{lem}{Lemma}[section]

\newtheorem{thm}[lem]{Theorem}
\newtheorem{cor}[lem]{Corollary}

\newtheorem{defn}[lem]{Definition}

\newtheorem{prop}[lem]{Proposition}

\newcommand{\bracket}[1]{\left(#1\right)}
\newcommand{\sqbracket}[1]{\left[#1\right]}

\makeatletter
\newcommand{\vast}{\bBigg@{4}}
\newcommand{\Vast}{\bBigg@{5}}
\makeatother

\newcommand{\ltwo}[1]{\left\|#1\right\|_2}

\newcommand{\norm}[1]{\| #1 \|}

\DeclarePairedDelimiterX{\infdivx}[2]{(}{)}{%
  #1\;\delimsize\|\;#2%
}

\newcommand*\samethanks[1][\value{footnote}]{\footnotemark[#1]}

%% file: abstract.tex
\begin{abstract}
Differentially Private Stochastic Gradient Descent (DP-SGD) forms a fundamental building block in many applications for learning over sensitive data. Two standard approaches, privacy amplification by subsampling, and privacy amplification by shuffling, permit adding lower noise in DP-SGD than via na\"{\i}ve schemes. A key assumption in both these approaches is that the elements in the data set can be uniformly sampled, or be uniformly permuted ---  constraints that may become prohibitive when the data is processed in a decentralized or distributed fashion. In this paper, we focus on conducting iterative methods like DP-SGD in the setting of federated learning (FL) wherein the data is distributed among many devices (clients). Our main contribution is the \emph{random check-in} distributed protocol, which crucially relies only on randomized participation decisions made locally and independently by each client. It has privacy/accuracy trade-offs similar to privacy amplification by subsampling/shuffling. However, our method does not require server-initiated communication, or even knowledge of the population size. To our knowledge, this is the first privacy amplification tailored for a distributed learning framework, and it may have broader applicability beyond FL. Along the way, we improve the privacy guarantees of amplification by shuffling and show that, in practical regimes, this improvement allows for similar privacy and utility using data from an order of magnitude fewer users.

\end{abstract}

%% file: intro.tex
\section{Introduction}
\label{sec:intro}

Modern mobile devices and web services benefit significantly from large-scale machine learning, often involving training on user (client) data. When such data is sensitive, steps must be taken to ensure privacy, and a formal guarantee of differential privacy (DP)~\cite{DMNS, ODO} is the gold standard.
For this reason, DP has been adopted by companies including Google~\cite{rappor,esa,esa++}, Apple~\cite{apple_report}, Microsoft~\cite{mic}, and LinkedIn~\cite{link}, as well as the US Census Bureau~\cite{census2}. 

Other privacy-enhancing techniques can be combined with DP to obtain additional benefits. 
In particular, cross-device federated learning (FL) \cite{FL1} allows model training while keeping client data decentralized (each participating device keeps its own local dataset, and only sends model updates or gradients to the coordinating server). 
However, existing approaches to combining FL and DP make a number of assumptions that are unrealistic in real-world  FL deployments such as \citep{bonawitz2019towards}.
To highlight these challenges, we must first review the state-of-the-art in centralized DP training, where differentially private stochastic gradient descent (DP-SGD) \cite{song2013stochastic, BST14, DP-DL} is ubiquitous. It achieves optimal error for convex problems \cite{BST14}, and can also be applied to non-convex problems, including deep learning, where the privacy amplification offered by randomly subsampling data to form batches is critical for obtaining meaningful DP guarantees \cite{KLNRS, BST14, DP-DL, BBG18, WangBK19}.

Attempts to combine FL and the above lines of DP research have been made previously; notably, \cite{MRTZ18, augenstein2019generative} extended the approach of \cite{DP-DL} to FL and user-level DP. However, these works and others in the area sidestep a critical issue: the DP guarantees require very specific sampling or shuffling schemes assuming, for example, that each client participates in each iteration with a fixed probability.
While possible in theory, such schemes are incompatible with the practical constraints and design goals of cross-device FL protocols \cite{bonawitz2019towards}; to quote \cite{FLO}, a comprehensive recent FL survey, \emph{``such a sampling procedure is nearly impossible in practice.''}\footnote{In cross-silo FL applications~\cite{FLO}, an enumerated set of addressable institutions or data-silos participate in FL, and so explicit server-mediated subsampling or shuffling using existing techniques may be feasible.} The fundamental challenge is that clients decide when they will be available for training and when they will check in to the server, and by design the server cannot index specific clients. In fact, it may not even know the size of the participating population. 

Our work targets these challenges. Our primary goal is to provide strong central DP guarantees for the final model released by FL-like protocols, under the assumption of a trusted\footnote{Notably, our guarantees are obtained by amplifying the privacy provided by local DP randomizers; we treat this use of local DP as an implementation detail in accomplishing the primary goal of central DP. As a byproduct, our approach offers (weaker) local DP guarantees even in the presence of an untrusted server.} orchestrating server.
This is accomplished by building upon recent work on amplification by shuffling \cite{soda-shuffling,dpmixnets, esa++,BPV20, privacy-blanket} and combining it with new analysis techniques targeting FL-specific challenges (e.g.,\ client-initiated communications, non-addressable global population, and constrained client availability).

\begin{figure}[ht]
	\centering
	\begin{subfigure}[b]{0.4\columnwidth}
        \centering
		\includegraphics[width=0.7\textwidth]{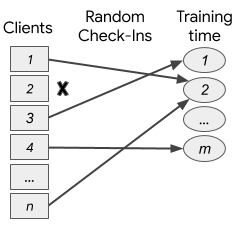}
		\caption{}
		\label{fig:client}
	\end{subfigure}
	\begin{subfigure}[b]{0.59\columnwidth}
        \centering
		\includegraphics[width=0.8\textwidth]{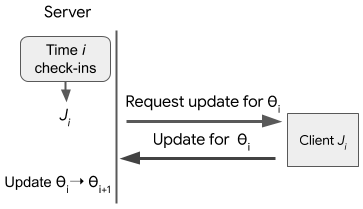}
		\caption{}
		\label{fig:server}
	\end{subfigure}
    \caption{A schematic of the Random Check-ins protocol with Fixed Windows (Section~\ref{sec:fixed_window}) for Distributed DP-SGD (Algorithm~\ref{alg:a_dist_new}).
    For the central DP guarantee, all solid arrows represent communication over privileged channels not accessible to any external adversary. 
    (a) $n$ clients performing random check-ins with a fixed window of $m$ time steps. `X' denotes  that the client randomly chose to abstain from participating. 
    (b) A time step at the server, where for training time $i \in [m]$, the server selects a client $j$ from those who checked-in for time $i$, requests an update for model $\theta_i$, and then updates the model to $\theta_{i+1}$ (or gradient accumulator if using minibatches).}
	\label{fig:rci_new}
\end{figure}

We propose the first privacy amplification analysis specifically tailored for distributed learning frameworks.
At the heart of our  result is a novel technique, called \emph{random check-in}, that relies only on randomness independently generated by each individual client participating in the training procedure.
We show that distributed learning protocols based on random check-ins can attain privacy gains similar to privacy amplification by subsampling/shuffling (see Table~\ref{table:comp} for a comparison), while requiring minimal coordination from the server.
While we restrict our exposition to distributed DP-SGD within the FL framework for clarity and concreteness (see Figure~\ref{fig:rci_new} for a schematic of one of our protocols), we note that the techniques used in our analyses are broadly applicable to any distributed iterative method and might be of interest in other applications\footnote{In particular, the Federated Averaging \cite{FL1} algorithm, which computes an update based on multiple local SGD steps rather than a single gradient, can immediately be plugged into our framework.}.

\paragraph{Contributions} The main contributions of this paper can be summarized as follows:
\begin{enumerate}[leftmargin=1em,topsep=0pt]
    \item We propose \emph{random check-ins}, the first privacy amplification technique for distributed systems with minimal server-side overhead. We also instantiate three distributed learning protocols that use random check-ins, each addressing different natural constraints that arise in applications.
    \item We provide formal privacy guarantees for our protocols, and show that random check-ins attain similar rates of privacy amplification as subsampling and shuffling while reducing the need for server-side orchestration.
    We also provide utility guarantees for one of our protocols in the convex case that match the optimal privacy/accuracy trade-offs for DP-SGD in the central setting \cite{BassilyFTT19}.
    \item As a byproduct of our analysis, we improve privacy amplification by shuffling \cite{soda-shuffling} on two fronts. For the case of $\epsilon_0$-DP local randomizers, we improve the dependency of the final central DP $\epsilon$ by a factor of $O(e^{0.5 \eps_0})$. Figure~\ref{fig:shuffling_comparison} provides a numerical comparison of the bound from \cite{soda-shuffling} with our bound; for typical parameter values this improvement allows us to provide similar privacy guarantees while reducing the number of required users by one order of magnitude. We also extend the analysis to the case of $(\epsilon_0, \delta_0)$-DP local randomizers, including Gaussian randomizers that are widely used in practice.

\begin{figure}
  \begin{minipage}[c]{0.45\textwidth}
\centering
    \includegraphics[width=0.85\textwidth]{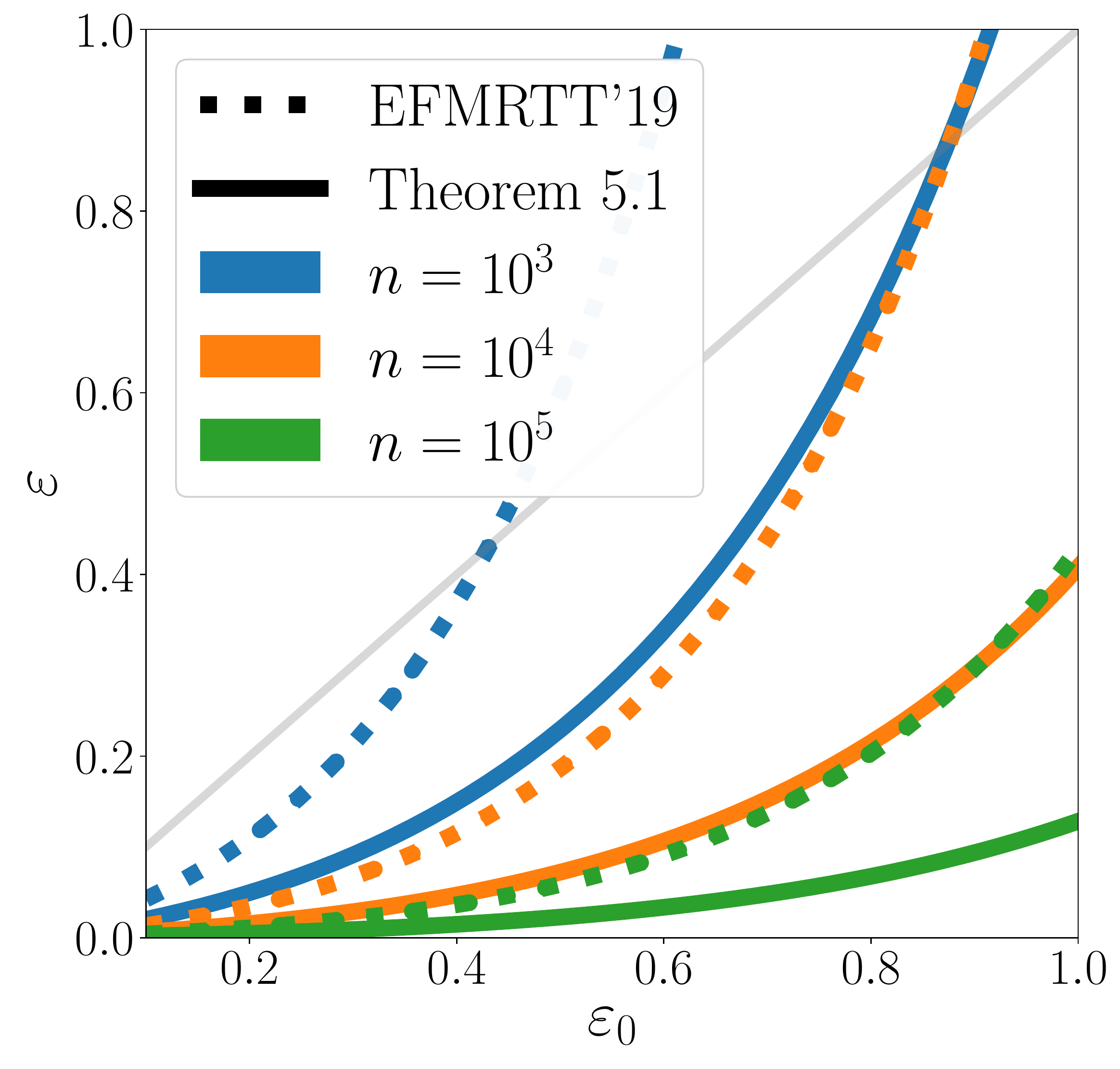}
  \end{minipage}
  \begin{minipage}[c]{0.54\textwidth}
    \caption{\small
       Values of $\eps$ (for $\delta = 10^{-6}$) after amplification by shuffling of $\eps_0$-DP local randomizers obtained from: Theorem~\ref{thm:ampl_shuff} (solid lines) and \cite[Theorem 7]{soda-shuffling} (dotted lines). The grey line represents the threshold of no amplification ($\eps = \eps_0$); after crossing the line amplification bounds become vacuous. Observe that our bounds with $n = 10^3$ and $n=10^4$ are similar to the bounds from \cite{soda-shuffling} with $n=10^4$ and $n=10^5$, respectively.
    } \label{fig:shuffling_comparison}
  \end{minipage}
\end{figure}

\end{enumerate}

\paragraph{Related work}
Our work considers the paradigm of federated learning as a stylized example throughout the paper. 
We refer the reader to \cite{FLO} for an excellent overview of the state-of-the-art in federated learning, along with a suite of interesting open problems.
There is a rich literature on studying differentially private ERM via DP-SGD  \cite{song2013stochastic, BST14, DP-DL, WLKCJN17, TAB19, pichapati2019adaclip}. However, constraints such as limited availability in distributed settings restrict direct applications of existing techniques. 
There is also a growing line of works on privacy amplification by shuffling~ \cite{esa, soda-shuffling,dpmixnets, BC19, privacy-blanket, BPV20,  esa++} that focus on various ways in which protocols can be designed using trusted shuffling primitives. 
Lastly, privacy amplification by iteration~\cite{FMTT18} is another recent advancement that can be applied in an iterative distributed setting, but it is limited to convex objectives.

%% file: prelims.tex
\section{Background and Problem Formulation}
\label{sec:back}

\paragraph{Differential Privacy} To formally introduce our notion of privacy, we first define neighboring data sets. We will refer to a pair of data sets
$D, D' \in \calD^n$ as neighbors if $D'$
can be obtained from
$D$ by modifying one sample $d_i \in D$ for some $i \in [n]$.

\begin{defn}[Differential privacy \cite{DMNS, ODO}] A randomized algorithm $\calA : \calD^n \to \calS$ is $(\eps,\delta)$-differentially private if, for any pair of neighboring data sets $D, D' \in \calD^n$, and for all events $S \subseteq \calS$ in the output range of $\calA$, we have 
$\Pr[\calA(D)\in S] \leq e^{\eps} \cdot \Pr[\calA(D')\in S] +\delta$.
\label{def:diffP}
\end{defn}

For meaningful \emph{central DP} guarantees (i.e., when $n > 1$), $\epsilon$ is assumed to be a small constant, and $\delta \ll 1/n$.
The case $\delta = 0$ is often referred to as \emph{pure DP} (in which case, we just write $\eps$-DP).
We shall also use the term \emph{approximate DP} when $\delta > 0$. 

Adaptive differentially private mechanisms occur naturally when constructing complex DP algorithms, for e.g., DP-SGD.
In addition to the dataset $D$, adaptive mechanisms also receive as input the output of other differentially private mechanisms.
Formally, we say that an adaptive mechanism $\calA : \calS' \times \calD^n \to \calS$ is $(\eps,\delta)$-DP if the mechanism $\calA(s',\bullet)$ is $(\eps,\delta)$-DP for every $s' \in \calS'$.

Specializing Definition~\ref{def:diffP} to the case $n = 1$ gives what we call a \emph{local randomizer}, which provides a \emph{local DP} guarantee. Local randomizers are the typical building blocks of local DP protocols where individuals privatize their data before sending it to an aggregator for analysis~\cite{KLNRS}.

\paragraph{Problem Setup} The distributed learning setup we consider in this paper involves $n$ clients, where each client $j \in [n]$ holds a data record\footnote{Each client is identified as a user. In a general FL setting, each $d_j$ can correspond to a local data set \cite{bonawitz2019towards}.} $d_j \in \calD$, $j \in [n]$, forming a distributed data set $D = (d_1, \ldots, d_n)$.
We assume a coordinating server wants to train the parameters $\theta \in \Theta$ of a model by using the dataset $D$ to perform stochastic gradient descent steps according to some loss function $\ell: \calD \times \Theta \rightarrow \mathbb{R}_+$.
The server's goal is to protect the privacy of all the individuals in $D$ by providing strong DP guarantees against an adversary that can observe the final trained model as well as all the intermediate model parameters.
We assume the server is trusted, all devices adhere to the prescribed protocol (i.e.,\ there are no malicious users), and all server-client communications are privileged (i.e., they cannot be detected or eavesdropped by an external adversary).

The server starts with model parameters $\theta_1$ and over a sequence of $m$ time slots produces a sequence of model parameters $\theta_2, \ldots, \theta_{m+1}$. 
Our random check-ins technique allows clients to independently decide when to offer their contributions for a model update.
If and when a client's contribution is accepted by the server, she uses the current parameters $\theta$ and her data $d$ to send a privatized gradient of the form $\calA_{ldp}(\nabla_{\theta} \ell(d, \theta))$ to the server, where $\calA_{ldp}$ is a DP local randomizer (e.g., performing gradient clipping and adding Gaussian noise \cite{DP-DL}).

Our results consider three different setups inspired by practical applications \cite{bonawitz2019towards}: 
(1) The server uses $m \ll n$ time slots, where at most one user's update is used in each slot, for a total of $m/b$ minibatch SGD iterations. 
It is assumed all $n$ users are available for the duration of the protocol, but the server does not have enough bandwidth to process updates from every user (Section~\ref{sec:fixed_window}); 
(2) The server uses $m \approx n/b$ time slots, and all $n$ users are available for the duration of the protocol (Section~\ref{sec:better_updates}). On average, $b$ users contribute updates to each time slot, and so, we take $m$ minibatch SGD steps; 
(3) As with (2), but each user is only available during a small window of time relative to the duration of the protocol (Section~\ref{sec:sliding_checkins}).

%% file: new_checkins.tex
\section{Distributed Learning with Random Check-Ins}
\label{sec:fixed_checkins}

This section presents the \emph{random check-ins} technique for privacy amplification in the context of distributed learning.
We formally define the random check-ins procedure, describe a fully distributed DP-SGD protocol with random check-ins, and analyze its privacy and utility guarantees.

\subsection{Random Check-Ins with a Fixed Window}\label{sec:fixed_window}

Consider the distributed learning setup described in Section~\ref{sec:back} where each client is willing to participate in the training procedure as long as their data remains private.
To boost the privacy guarantees provided by the local randomizer $\calA_{ldp}$, we will let clients volunteer their updates at a \emph{random} time slot of their choosing.
This randomization has a similar effect on the uncertainty about the use of an individual's data on a particular update as the one provided by uniform subsampling or shuffling.
We formalize this concept using the notion of random check-in, which can be informally expressed as a client in a distributed iterative learning framework randomizing their instant of participation, and determining with some probability whether to participate in the process at all.

\begin{defn}[Random check-in]
Let $\calA$ be a distributed learning protocol with $m$ check-in time slots. For a set $R_j \subseteq [m]$ and probability $p_j \in [0, 1]$, client $j$ performs an $(R_j, p_j)$-check-in in the protocol if with probability $p_j$ she requests the server to participate in $\calA$ at time step $I \xleftarrow{u.a.r.} R_j$, and otherwise abstains from participating.
If $p_j = 1$, we alternatively denote it as an $R_j$-check-in.
\end{defn}

Our first distributed learning protocol based on random check-ins is presented in Algorithm~\ref{alg:a_dist_new}.
Client $j$ independently decides in which of the possible time steps (if any) she is willing to participate by performing an $(R_j, p_j)$-check-in.
We set $R_j = [m]$ for all $j \in [n]$, and assume\footnote{We make this assumption only for utility; the privacy guarantees are independent of this assumption.} all $n$ clients are available throughout the duration of the protocol.
On the server side, at each time step $i \in [m]$, a random client $J_i$ among all the ones that checked-in at time $i$ is queried: this client receives the current model $\theta_i$, locally computes a gradient update $\nabla_{\theta} \ell(d_{J_i}, \theta_i)$ using their data $d_{J_i}$, and returns to the server a privatized version of the gradient obtained using a local randomizer $\calA_{ldp}$.
Clients checked-in at time $i$ that are not selected do not participate in the training procedure.
If at time $i$ no client is available, the server adds a ``dummy" gradient to update the model.

\begin{figure}
\begin{small}
\centering
\begin{minipage}[t]{0.49\textwidth} 
\begin{center}
\begin{algorithmic}
\SUB{Server-side protocol:}
\STATE \emph{parameters:} local randomizer $\calA_{ldp}$, number of steps $m$ \\[0.5em]
\STATE Initialize model $\theta_1 \in \Theta$
\STATE Initialize gradient accumulator $g_1 \gets 0^p$
\FOR{$i \in [m]$}
\STATE $S_i \leftarrow \{j : \text{User}(j) \text{ checked-in at time } i \}$
\IF{$S_i$ is empty}
\STATE $\tilde{g}_i \leftarrow \calA_{ldp}\left(0^p \right) $ \hfill // Dummy gradient
\ELSE
\STATE Sample $J_i \xleftarrow{u.a.r.} S_i$
\STATE Request User$(J_i)$ for update to model $\theta_{i}$
\STATE Receive $\tilde{g}_i $ from User$(J_i)$
\ENDIF
\STATE $(\theta_{i+1}, g_{i+1}) \gets \text{ModelUpdate}(\theta_i, g_i + \tilde{g}_i, i)$
\STATE Output $\theta_{i+1}$
\ENDFOR
\end{algorithmic}
\end{center} 
\vfill
\end{minipage}
\hfill
\begin{minipage}[t]{0.49\textwidth}
\begin{center}
\begin{algorithmic}
\SUB{Client-side protocol for User$(j)$:}
\STATE \emph{parameters:} check-in window $R_j$, check-in probability $p_j$, loss function $\ell$, local randomizer $\calA_{ldp}$
\STATE \emph{private inputs:} datapoint $d_j \in \calD$ \\[0.5em]
\IF{a $p_j$-biased coin returns heads}
\STATE Check-in with the server at time $I \xleftarrow{u.a.r.} R_j$
\IF{receive request for update to model $\theta_{I}$}
\STATE $\tilde{g}_{I} \leftarrow \calA_{ldp}(\nabla_{\theta}\ell(d_j, \theta_{I}))$
\STATE Send $\tilde{g}_{I}$ to server
\ENDIF
\ENDIF
\end{algorithmic}
\end{center}
\vspace*{-.5em}
\begin{center}
\begin{algorithmic}
\SUB{ModelUpdate$(\theta, g, i)$:}
\STATE \emph{parameters:} batch size $b$, learning rate $\eta$ \\[0.5em]
\IF{$i\mod b = 0$}
\RETURN $\left(\theta - \frac{\eta}{b 
} g, 0^p\right)$ \hfill // Gradient descent step 
\ELSE
\RETURN $(\theta, g)$  \hfill // Skip update
\ENDIF
\end{algorithmic}
\end{center}
\vfill
\end{minipage}
\end{small}
\rule{\textwidth}{0.4pt} 
\mycaptionof{algorithm}{$\calA_{fix}$ -- Distributed DP-SGD with random check-ins (fixed window).}
\label{alg:a_dist_new}
\end{figure}

\subsection{Privacy Analysis}

From a privacy standpoint, Algorithm~\ref{alg:a_dist_new} shares an important pattern with DP-SGD: each model update uses noisy gradients obtained from a random subset of the population.
However, there exist two key factors that make the privacy analysis of our protocol more challenging than the existing analysis based on subsampling and shuffling.
First, unlike in the case of uniform sampling where the randomness in each update is independent, here there is a correlation induced by the fact that clients that check-in into one step cannot check-in into a different step.
Second, in shuffling there is also a similar correlation between updates, but there we can ensure each update uses the same number of datapoints, while here the server does not control the number of clients that will check-in into each individual step.
Nonetheless, the following result shows that random check-ins provides a factor of privacy amplification comparable to these techniques.

\begin{thm}[Amplification via random check-ins into a fixed window]
Suppose $\calA_{ldp}$ is an $\eps_0$-DP local randomizer. 
Let $\calA_{fix}: \calD^{n} \rightarrow \Theta^{m}$ be the protocol from Algorithm~\ref{alg:a_dist_new} with check-in probability $p_j=p_0$ and check-in window $R_j = [m]$ for each client $j \in [n]$.
For any $\delta \in (0,1)$, algorithm $\calA_{fix}$ is $\left(\eps, \delta\right)$-DP with $\eps = p_0(e^{\eps_0} - 1)\sqrt{\frac{2 e^{\eps_0} \log{(1/\delta)}}{m}} + \frac{p_0^2 e^{\eps_0} (e^{\eps_0} - 1)^2}{2m}$. In particular, for $\eps_0 \leq 1$ and $\delta \leq 1/100$, we get $\eps \leq 7p_0\eps_0\sqrt{\frac{\log(1/\delta)}{m}}$.
Furthermore, if $\calA_{ldp}$ is $(\eps_0,\delta_0)$-DP with $\delta_0 \leq \frac{(1 - e^{-\eps_0}) \delta_1}{4 e^{\eps_0} \left(2 + \frac{\ln(2/\delta_1)}{\ln(1/(1-e^{-5\eps_0}))}\right)}$, then $\calA_{fix}$ is $(\eps',\delta')$-DP with $\eps' = \frac{p_0^2 e^{8\eps_0} (e^{8\eps_0} - 1)^2}{2m} + p_0(e^{8\eps_0} - 1)\sqrt{\frac{2 e^{8\eps_0} \log{(1/\delta)}}{m}}$ and $\delta' = \delta + m (e^{\eps'}+1) \delta_1$.
\label{thm:ampl_dist+}
\end{thm}

\paragraph{Remark 1} We can always increase privacy in the above statement by decreasing $p_0$. However, this will also increase the number of dummy updates, which suggests choosing $p_0 = \Theta(m/n)$. With such a choice, we obtain an amplification factor of $\sqrt{m}/n$. Critically, however, exact knowledge of the population size is \emph{not} required to have a precise DP guarantee above. 

\paragraph{Remark 2} At first look, the amplification factor of $\sqrt{m}/n$ may appear stronger than the typical $1/\sqrt{n}$ factor obtained via uniform subsampling/shuffling. Note that one run of our technique provides $m$ updates (as opposed to $n$ updates via the other methods). When the server has sufficient capacity, we can set $m = n$ to recover a $1/\sqrt{n}$ amplification. The primary advantage of our approach is that we can benefit from amplification in terms of  $n$ even if only a much smaller number of updates are actually processed. We can also extend our approach to recover the $1/\sqrt{n}$ amplification even when the server is rate limited ($p_0 = m/n$), by repeating the protocol $\calA_{fix}$ adaptively $n/m$ times to get Corollary~\ref{cor:res} from Theorem~\ref{thm:ampl_dist+} and applying advanced composition for DP~\cite{dwork2014algorithmic}.
\begin{cor}
For algorithm $\calA_{fix}: \calD^{n} \rightarrow \Theta^{m}$ described in Theorem~\ref{thm:ampl_dist+}, suppose $\calA_{ldp}$ is an $\eps_0$-DP local randomizer s.t. 
$\eps_0 \leq \frac{2 \log{\bracket{n/8\sqrt{m}}}}{3}$, and 
$n \geq (e^{\eps_0} - 1)^2e^{\eps_0}\sqrt{m}\log\bracket{1/\beta}$. Setting $p_0=\frac{m}{n}$, and running $\frac{n}{m}$ repetitions of $\calA_{fix}$ results in a total of $n$ updates, along with an overall central $(\eps, \delta)$-DP guarantee with $\eps = \tilde{O}\left( e^{1.5\eps_0} / \sqrt{n}\right)$ and $\delta \in (0, 1)$, where $\tilde{O}(\cdot)$ hides polylog factors in $1/ \beta$ and $1/ \delta$.
\label{cor:res}
\end{cor}

\paragraph{Comparison to Existing Privacy Amplification Techniques}

Table~\ref{table:comp} provides a comparison of the bound in Corollary~\ref{cor:res} to other existing techniques, for performing one epoch of training (i.e., use one update from each client). Note that for this comparison, we assume that $\eps_0 > 1$, since for $\eps_0 \leq 1$ all the shown amplification bounds can be written as $O\bracket{\eps_0 / \sqrt{n}}$.
``None" denotes a na\"{\i}ve scheme (with no privacy amplification) where each client is used exactly once in any arbitrary order.
Also, note that in general, the guarantees via privacy amplification by subsampling/shuffling apply only under the assumption of complete participation availability\footnote{By a complete participation availability for a client, we mean that the client should be available to participate when requested by the server for any time step(s) of training.} of each client. Thus, they define the upper limits of achieving such amplifications.  
 Also, note that even though the bound in Corollary~\ref{cor:res} appears better than amplification via shuffling, 
 our technique does include dummy updates which do not occur in the other techniques. For linear optimization problems, it is easy to see that our technique will add a factor of $e$ more noise as compared to the other two privacy amplification techniques at the same privacy level.

\begin{table}[ht]
\begin{minipage}[c]{.6\textwidth}
{\renewcommand{\arraystretch}{1.2}
    \centering
    \begin{tabular}{| c | c |}
    \hline
    \textbf{Source of Privacy Amplification} & \textbf{$\eps$ for Central DP} \\
    \hline
    None~\cite{DJW13, smith2017interaction} & $\eps_0$ \\
    \hline
    Uniform subsampling~\cite{KLNRS, BST14, DP-DL} & $O\left(e^{\eps_0} / \sqrt{n}\right)$ \\
    \hline
    Shuffling~\cite{soda-shuffling} & $O\left(e^{3\eps_0} / \sqrt{n}\right)$ \\
    \hline
    Shuffling (Theorem~\ref{thm:ampl_shuff}, This paper) & $O\left(e^{2.5\eps_0} / \sqrt{n}\right)$ \\
    \hline
    Random check-ins with a fixed window  & $O\left( e^{1.5\eps_0} / \sqrt{n}\right)$ \\
    (Theorem~\ref{thm:ampl_dist+}, This paper) & \\
    \hline
    \end{tabular}
}
\end{minipage}
\hfill
\begin{minipage}[c]{.35\textwidth}
\caption{\small Comparison with existing amplification techniques for a data set of size $n$, running $n$ iterations of DP-SGD with batch size of 1 and $\eps_0$-DP local randomizers. For ease of exposition, we assume $(e^{\eps_0} - 1) \approx \eps_0$, and hide polylog factors in $n$ and $1/ \delta$. } 
    \label{table:comp}
\end{minipage}
\end{table}

\paragraph{Proof Sketch for Theorem~\ref{thm:ampl_dist+}}
Here, we provide a summary of the argument\footnote{Full proofs for every result in the paper are provided in Appendix~\ref{sec:proofs}.} used to prove Theorem~\ref{thm:ampl_dist+} in the case $\delta_0 = 0$.
First, note that  it is enough to argue about the privacy of the sequence of noisy gradients $\tilde{g}_{1:m}$ by post-processing.
Also, the role each client plays in the protocol is symmetric, so w.l.o.g.\ we can consider two datasets $D, D'$ differing in the first position.
Next, we imagine that the last $n-1$ clients make the same random check-in choices in $\calA_{fix}(D)$ and $\calA_{fix}(D')$.
Letting $c_i$ denote the number of such clients that check-in into step $i \in [n]$, we model these choices by a pair of sequences $F = (\bar{d}_{1:m}, w_{1:m})$ where $\bar{d}_i \in \calD \cup \{\bot\}$ is the data record of an arbitrary client who checked-in into step $i$ (with $\bot$ representing a ``dummy'' data record if no client checked-in), and $w_i = 1/(c_i + 1)$ represents the probability that client $1$'s data will be picked to participate in the protocol at step $i$ if she checks-in in step $i$.
Conditioned on these choices, the noisy gradients $\tilde{g}_{1:m}$ produced by $\calA_{fix}(D)$ can be obtained by: (1) initializing a dataset $\tilde{D} = (\bar{d}_{1:m})$; (2) sampling $I \xleftarrow{u.a.r.} [m]$, and replacing $\bar{d}_I$ with $d_1$ in $\tilde{D}$ w.p. $p_0 w_I$; (3) producing the outputs $\tilde{g}_{1:m}$ by applying a sequence of $\eps_0$-DP adaptive local randomizers to $\tilde{D} = (\tilde{d}_{1:m})$ by setting $\tilde{g}_i = \calA^{(i)}(\tilde{d}_i, \tilde{g}_{1:i-1})$.
Here each of the $\calA^{(i)}$ uses all past gradients to compute the model $\theta_i$ and return $\tilde{g}_i = \calA_{ldp}(\nabla_{\theta} \ell(\tilde{d}_i, \theta_i))$.

The final step involves a variant of the amplification by swapping technique \cite[Theorem 8]{soda-shuffling} which we call amplification by probable replacement.
The key idea is to reformulate the composition of the $\calA^{(i)}$ applied to the random dataset $\tilde{D}$, to a composition of mechanisms of the form $\tilde{g}_i = \calB^{(i)}(d_1, F, \tilde{g}_{1:i-1})$.
Mechanism $\calB^{(i)}$ uses the gradient history to compute $q_i = \Pr[I = i | \tilde{g}_{1:i-1}]$ and
returns $\calA^{(i)}(d_1, \tilde{g}_{1:i-1})$ with probability $p_0 w_i q_i$, and $\calA^{(i)}(\bar{d}_i, \tilde{g}_{1:i-1})$ otherwise.
Note that before the process begins, we have $\Pr[I = i] = 1/m$ for every $i$; our analysis shows that the posterior probability after observing the first $i - 1$ gradients is not too far from the prior: $q_i \leq \frac{e^{\eps_0}}{m e^{\eps_0} - (e^{\eps_0}-1)(i-1)}$.
The desired bound is then obtained by using the overlapping mixtures technique \cite{BBG18} to show that $\calB^{(i)}$ is $\log(1 + p_0 q_i (e^{\eps_0}-1))$-DP with respect to changes on $d_1$, and heterogeneous advanced composition \cite{KOV17} to compute the final $\eps$ of composing the $\calB^{(i)}$ adaptively.

\subsection{Utility Analysis}

\begin{prop}[Dummy updates in random check-ins with a fixed window]
For algorithm $\calA_{fix}: \calD^{n} \rightarrow \Theta^{m}$ described in Theorem~\ref{thm:ampl_dist+}, the expected number of dummy updates performed by the server is at most $ \left(m  \left(1 - \frac{p_0}{m}\right)^n\right)$. For  $c > 0$ if $p_0 = \frac{cm}{n}$, we get at most $\frac{m}{e^c}$ expected dummy updates. 
\label{prop:dummy_dist+}
\end{prop}

\paragraph{Utility for Convex ERMs} We now instantiate our amplification theorem (Theorem~\ref{thm:ampl_dist+}) in the context of differentially private empirical risk minimization (ERM). 
For convex ERMs, we will show that DP-SGD~\cite{song2013stochastic,BST14,DP-DL} in conjunction with our privacy amplification theorem (Theorem~\ref{thm:ampl_dist+}) is capable of achieving the optimal privacy/accuracy trade-offs~\cite{BST14}.

\begin{thm}[Utility guarantee]
Suppose in algorithm $\calA_{fix}: \calD^{n} \rightarrow \Theta^{m}$ described in Theorem~\ref{thm:ampl_dist+} 
the loss $\ell:\calD\times\Theta \rightarrow \mathbb{R}_+$ is $L$-Lipschitz and convex in its second parameter and the model space $\Theta$ has dimension $p$ and diameter $R$, i.e., $\sup_{\theta, \theta' \in \Theta}\|\theta - \theta'\|\leq R$. Furthermore, let $\mathscr{D}$ be a distribution on $\calD$, define the population risk $\mathscr{L}(\mathscr{D}; \theta) = \mathbb{E}_{d \sim \mathscr{D}}\left[ \ell(d; \theta) \right]$, and let $\theta^* = \arg\min_{\theta \in \Theta}\mathscr{L}(\mathscr{D}; \theta)$. 
If $\calA_{ldp}$ is a local randomizer that adds Gaussian noise with variance $\sigma^2$,
and the learning rate for a model update at step $i \in [m]$ is set to be $\eta_i = \frac{R \left(1 - 2e^{-np_0 / m}\right)}{\sqrt{\bracket{p\sigma^2 + L^2}i}}$,
then the output $\theta_m$ of $\calA_{fix}(D)$ on a dataset $D$ containing $n$ i.i.d.\ samples from $\mathscr{D}$ satisfies\footnote{Here, $\widetilde{O}$ hides a polylog factor in $m$.}  
$$\mathbb{E}_{D,\theta_m}\left[\mathscr{L}(\mathscr{D}; \theta_m) \right] - \mathscr{L}(\mathscr{D}; \theta^*) = \widetilde{O}\left(\frac{\sqrt{p\sigma^2 + L^2} \cdot R}{\left(1 - 2e^{-np_0 / m}\right)\sqrt{m}} \right).$$
\label{thm:util_new}
\end{thm}

\paragraph{Remark 3} Note that as $m \to n$,
it is easy to see for $p_0 = \Omega\left(\frac{m}{n}\right)$ that Theorem~\ref{thm:util_new} achieves the optimal population risk trade-off \cite{BST14, BassilyFTT19}. 

\section{Variations: Thrifty Updates, and Sliding Windows}
\label{sec:variations}

This section presents two variants of the main protocol from the previous section.
The first variant makes a better use of the updates provided by each user at the expense of a small increase in the privacy cost.
The second variant allows users to check-in into a sliding window to model the case where different users might be available during different time windows.

\subsection{Leveraging Updates from Multiple Users}
\label{sec:better_updates}

\begin{figure}
\centering
\begin{minipage}[h]{0.45\textwidth}
\begin{center}
\begin{algorithmic}
\SUB{Server-side protocol:}
   \STATE \emph{parameters:} total update steps $m$
   \STATE
   \STATE Initialize model $\theta_1 \in \mathbb{R}^p$
   \FOR{$i \in [m]$}
        \STATE $S_i \leftarrow \{j : \text{User}(j) \text{ checks-in for index } i \}$
        \IF{$S_i$ is empty}
            \STATE $\theta_{i+1} \gets \theta_i$
        \ELSE
            \STATE $\tilde{g}_i \gets 0$
            \FOR{$j \in S_i$}
                \STATE Request User$(j)$ for update to model $\theta_{i}$
                \STATE Receive $\tilde{g}_{i,j}$ from User$(j)$
                \STATE $\tilde{g}_i \gets \tilde{g}_i + \tilde{g}_{i,j}$
            \ENDFOR
            \STATE $\theta_{i+1} \leftarrow \theta_i -  \frac{\eta}{|S_i|} \tilde{g}_i  $
        \ENDIF
        \STATE Output $\theta_{i+1}$
   \ENDFOR
\end{algorithmic}
\end{center}
\rule{\textwidth}{0.4pt}
\end{minipage}
\mycaptionof{algorithm}{$\calA_{avg}$ - Distributed DP-SGD with random check-ins (averaged updates).}
\label{alg:a_dist_avg}
\end{figure}

Now, we present a variant of Algorithm~\ref{alg:a_dist_new} which, at the expense of a mild increase in the privacy cost, removes the need for dummy updates, and for discarding all but one of the clients checked-in at every time step.
The server-side protocol of this version is given in Algorithm~\ref{alg:a_dist_avg} (the client-side protocol is identical as Algorithm~\ref{alg:a_dist_new}).
Note that here, if no client checked-in at some step $i \in [m]$, the server simply skips the update.
Furthermore, if at some step  multiple clients  checked in, the server requests gradients from all the clients, and performs a model update using the average of the submitted noisy gradients.

These changes have the obvious advantage of reducing the noise in the model coming from dummy updates, and increasing the algorithm's data efficiency by utilizing gradients provided by all available clients.
The corresponding privacy analysis becomes more challenging because (1) the adversary gains information about the time steps where no clients checked-in, and (2) the server uses the potentially non-private count $|S_i|$ of clients checked-in at time $i$ when performing the model update.
Nonetheless, we show that the privacy guarantees of Algorithm~\ref{alg:a_dist_avg} are similar to those of Algorithm~\ref{alg:a_dist_new} with an additional $O(e^{3 \eps_0/2})$ factor, and the restriction of non-collusion among the participating clients. 
For simplicity, we only analyze the case where each client has check-in probability $p_j = 1$.

\begin{thm}[Amplification via random check-ins with averaged updates]
Suppose $\calA_{ldp}$ is an $\eps_0$-DP local randomizer. 
Let $\calA_{avg}: \calD^{n} \rightarrow \Theta^{m}$ be the protocol from Algorithm~\ref{alg:a_dist_avg} performing $m$ averaged model updates with check-in probability $p_j=1$ and check-in window $R_j = [m]$ for each user $j \in [n]$.
Algorithm $\calA_{avg}$ is $\left(\eps, \delta + \delta_2 \right)$-DP with
\begin{align*}
    \eps =
    \frac{e^{4 \eps_0} (e^{\eps_0}-1)^2 \eps_1^2}{2}
    +
    e^{2 \eps_0} (e^{\eps_0}-1) \eps_1 \sqrt{2 \log(1/\delta)}
    \enspace,
\end{align*}
where $\eps_1 = \sqrt{\frac{1}{n} + \frac{1}{m}} + \sqrt{\frac{\log(1/\delta_2)}{n}}$.
In particular, for $\eps_0 \leq 1$ we get $\eps = O(\eps_0 / \sqrt{m})$.
Furthermore, if $\calA_{ldp}$ is $(\eps_0,\delta_0)$-DP with $\delta_0 \leq \frac{(1 - e^{-\eps_0}) \delta_1}{4 e^{\eps_0} \left(2 + \frac{\ln(2/\delta_1)}{\ln(1/(1-e^{-5\eps_0}))}\right)}$, then $\calA_{avg}$ is $(\eps', \delta')$-DP with $\eps' = \frac{e^{32 \eps_0} (e^{8\eps_0}-1)^2 \eps_1^2}{2} + e^{16 \eps_0} (e^{8\eps_0}-1) \eps_1 \sqrt{2 \log(1/\delta)}$ and $\delta' = \delta + \delta_2 + m (e^{\eps'} + 1) \delta_1$.
\label{thm:ampl_avg}
\end{thm}

Next, we provide a utility guarantee for $\calA_{avg}$ in terms of the excess population risk for convex ERMs (similar to Theorem~\ref{thm:util_new}).

\begin{thm}[Utility guarantee]
Suppose in algorithm $\calA_{avg}: \calD^{n} \rightarrow \Theta^{m}$ described in Theorem~\ref{thm:ampl_avg} 
the loss $\ell:\calD\times\Theta \rightarrow \mathbb{R}_+$ is $L$-Lipschitz and convex in its second parameter and the model space $\Theta$ has dimension $p$ and diameter $R$, i.e., $\sup_{\theta, \theta' \in \Theta}\|\theta - \theta'\|\leq R$. Furthermore, let $\mathscr{D}$ be a distribution on $\calD$, define the population risk $\mathscr{L}(\mathscr{D}; \theta) = \mathbb{E}_{d \sim \mathscr{D}}\left[ \ell(d; \theta) \right]$, and let $\theta^* = \arg\min_{\theta \in \Theta}\mathscr{L}(\mathscr{D}; \theta)$. 
If $\calA_{ldp}$ is a local randomizer that adds Gaussian noise with variance $\sigma^2$,
and the learning rate for a model update at step $i \in [m]$ is set to be $\eta_i = \frac{R\sqrt{n}}{\sqrt{\bracket{mp \sigma^2 + nL^2}i}}$,
then the output $\theta_m$ of $\calA_{avg}(D)$ on a dataset $D$ containing $n$ i.i.d.\ samples from $\mathscr{D}$ satisfies
$$\mathbb{E}_{D,\theta_m}\left[\mathscr{L}(\mathscr{D}; \theta_m) \right] - \mathscr{L}(\mathscr{D}; \theta^*) = \widetilde{O}\left(\frac{R\sqrt{mp \sigma^2 + nL^2}}{\sqrt{mn}} \right).$$
Furthermore, if the loss $\ell$ is $\beta$-smooth in its second parameter and we set the step-size $\eta_i = \frac{R \sqrt{n}}{\beta R \sqrt{n} + m \sqrt{L^2 + p \sigma^2}}$, then we have
$$\mathbb{E}_{D,\theta_1, \ldots, \theta_m}\left[\mathscr{L}\left(\mathscr{D}; \frac{1}{m} \sum_{i = 1}^m \theta_i\right) \right] - \mathscr{L}(\mathscr{D}; \theta^*) = \widetilde{O}\left(R \sqrt{\frac{L^2 + p \sigma^2}{n}} + \frac{\beta R^2}{m} \right).$$
\label{thm:util_new2}
\end{thm}

\mypar{Comparison to Algorithm~\ref{alg:a_dist_new} in Section~\ref{sec:fixed_checkins}}
Recall that in $\calA_{fix}$ we can achieve a small fixed $\eps$ by taking $p_0 = m/n$ and $\sigma = \tilde{O}(\frac{p_0}{\eps \sqrt{m}})$, in which case the excess risk bound in Theorem~\ref{thm:util_new} becomes $\tilde{O}\left(\sqrt{\frac{L^2}{m} + \frac{p}{\eps^2 n^2}} \right)$.
On the other hand, in $\calA_{avg}$ we can obtain a fixed small $\eps$ by taking $\sigma = \tilde{O}\bracket{\frac{1}{\eps \sqrt{m}}}$. In this case the excess risks in Theorem~\ref{thm:util_new2} are bounded by $\tilde{O}\bracket{\sqrt{\frac{L^2}{m} + \frac{p}{\eps^2 n m}}}$ in the convex case, and by $\tilde{O}\bracket{\sqrt{\frac{L^2}{n} + \frac{p}{\eps^2 n m}} + \frac{1}{m}}$ in the convex and smooth case. 
Thus, we observe that all the bounds recover the optimal population risk trade-offs from \cite{BST14, BassilyFTT19} as $m \to n$, and for $m \ll n$ and non-smooth loss $\calA_{avg}$ provides a better trade-off than $\calA_{fix}$, while on smooth losses $\calA_{avg}$ and $\calA_{fix}$ are incomparable.
Note that $\calA_{fix}$ (with $b =1$) will not attain a better bound on smooth losses because each update is based on a single data-point. Setting $b > 1$ will reduce the number of updates to $m/b$ for $\calA_{fix}$, whereas to get an excess risk bound for $\calA_{fix}$ for smooth losses where more than one data point is sampled at each time step will require extending the privacy analysis to incorporate the change, which is beyond the scope of this paper.

\subsection{Random Check-Ins with a Sliding Window}
\label{sec:sliding_checkins}

The second variant we consider removes the need for all clients to be available throughout the training period.
Instead, we assume that the training period comprises of $n$ time steps, and each client $j \in [n]$ is only available during a   window of $m$ time steps.
Clients perform a random check-in to provide the server with an update during their window of availability.
For simplicity, we assume clients wake up in order, one every time step, so client $j\in[n]$ will perform a random check-in within the window $R_j = \{j, \ldots, j+m-1\}$.
The server will perform $n - m + 1$ updates starting at time $m$ to provide a warm-up period where the first $m$ clients perform their random check-ins.

\begin{thm}[Amplification via random check-ins with sliding windows]
Suppose $\calA_{ldp}$ is an $\eps_0$-DP local randomizer. 
Let $\calA_{sldw}: \calD^{n} \rightarrow \Theta^{n-m+1}$ be the distributed algorithm performing $n$ model updates with check-in probability $p_j=1$ and check-in window $R_j = \{j,\ldots,j+m-1\}$ for each user $j \in [n]$.
For any $m \in [n]$, algorithm $\calA_{sldw}$ is $\left(\eps, \delta\right)$-DP with $\eps = \frac{e^{\eps_0} (e^{\eps_0} - 1)^2}{2m} + (e^{\eps_0} - 1)\sqrt{\frac{2 e^{\eps_0} \log{(1/\delta)}}{m}}$. For $\eps_0 \leq 1$ and $\delta \leq 1/100$, we get $\eps \leq 7\eps_0\sqrt{\frac{\log(1/\delta)}{m}}$.
Furthermore, if $\calA_{ldp}$ is $(\eps_0,\delta_0)$-DP with $\delta_0 \leq \frac{(1 - e^{-\eps_0}) \delta_1}{4 e^{\eps_0} \left(2 + \frac{\ln(2/\delta_1)}{\ln(1/(1-e^{-5\eps_0}))}\right)}$, then $\calA_{sldw}$ is $(\eps',\delta')$-DP with $\eps' = \frac{e^{8 \eps_0} (e^{8 \eps_0} - 1)^2}{2m} + (e^{8\eps_0} - 1)\sqrt{\frac{2 e^{8\eps_0} \log{(1/\delta)}}{m}}$ and $\delta' = \delta + m (e^{\eps'}+1) \delta_1$.
\label{thm:ampl_dist}
\end{thm}

\paragraph{Remark 4} We can always increase privacy in the statement above by increasing $m$.
However, that also increases the number of clients who do not participate in training because their scheduled check-in time is before the process begins, or after it terminates. Moreover, the number of empty slots where the server introduces dummy updates will also increase, which we would want to minimize for good accuracy. Thus,
$m$ introduces a trade-off between accuracy and privacy.

\begin{prop}[Dummy updates in random check-ins with sliding windows]
For algorithm $\calA_{sldw}: \calD^{n} \rightarrow \Theta^{n-m+1}$ described in Theorem~\ref{thm:ampl_dist}, the expected number of dummy gradient updates performed by the server is at most $(n - m + 1)/e$. 
\label{prop:dummy_dist}
\end{prop}

%% file: shuffling.tex
\section{Improvements to Amplification via Shuffling}
\label{sec:shuffling}

Here, we provide an improvement on privacy amplification by shuffling.
This is obtained using two technical lemmas 
\ifsupp
(Lemmas~\ref{lem:pr} and~\ref{lem:het})
\else
(deferred to the supplementary material) 
\fi
to tighten the analysis of amplification by swapping, a central component in the analysis of amplification by shuffling given in \cite{soda-shuffling}.

\begin{thm}[Amplification via Shuffling]
Let $\calA^{(i)}: \mathcal{S}^{(1)} \times \cdots \times \mathcal{S}^{(i - 1)} \times \calD \rightarrow \mathcal{S}^{(i)}$, $i \in [n]$, be a sequence of adaptive $\eps_0$-DP local randomizers.
Let $\calA_{sl}: \calD^{n} \rightarrow \mathcal{S}^{(1)} \times \cdots \times \mathcal{S}^{(n)}$ be the algorithm that given a dataset $D = (d_1, \ldots, d_n) \in \calD^{n}$ samples a uniform random permutation $\pi$ over $[n]$, sequentially computes $s_i = \calA^{(i)}(s_{1:i-1}, d_{\pi(i)})$ and outputs $s_{1:n}$.
For any $\delta \in (0,1)$, algorithm $\calA_{sl}$ satisfies $\left(\eps, \delta\right)$-DP with $\eps = \frac{e^{3\eps_0} (e^{\eps_0} - 1)^2}{2n} + e^{3\eps_0/2}(e^{\eps_0} - 1)\sqrt{\frac{2\log{(1/\delta)}}{n}}$.
Furthermore, if $\calA^{(i)}$, $i \in [n]$, is $(\eps_0, \delta_0)$-DP with $\delta_0 \leq \frac{(1 - e^{-\eps_0}) \delta_1}{4 e^{\eps_0} \left(2 + \frac{\ln(2/\delta_1)}{\ln(1/(1-e^{-5\eps_0}))}\right)}$, then $\calA_{sl}$ satisfies $(\eps', \delta')$-DP with $\eps' = \frac{e^{24\eps_0} (e^{8\eps_0} - 1)^2}{2n} + e^{12\eps_0}(e^{8\eps_0} - 1)\sqrt{\frac{2\log{(1/\delta)}}{n}}$ and $\delta' = \delta + n (e^{\eps'}+1) \delta_1$.
\label{thm:ampl_shuff}
\end{thm}

For comparison, the guarantee in \cite[Theorem 7]{soda-shuffling} in the case $\delta_0 = 0$ results in $$\eps = 2e^{2\eps_0} (e^{\eps_0} - 1) (e^{\frac{2\exp(2\eps_0) (e^{\eps_0} - 1)}{n}} - 1) + 2e^{2\eps_0}(e^{\eps_0} - 1)\sqrt{\frac{2  \log{(1/\delta)}}{n}}.$$

%% file: conclusion.tex
\section{Conclusion}
\label{sec:conc}

Our work highlights the fact that proving DP guarantees for distributed or decentralized systems can be substantially more challenging than for centralized systems, because in a distributed setting it becomes much harder to precisely control and characterize the randomness in the system, and this precise characterization and control of randomness is at the heart of DP guarantees.
 Specifically, production FL systems do not satisfy the assumptions that are typically made under state-of-the-art privacy accounting schemes, such as privacy amplification via subsampling. Without such accounting schemes, service providers cannot provide DP statements with small $\epsilon$'s. This work, though largely theoretical in nature, proposes a method shaped by the practical constraints of distributed systems that allows for rigorous privacy statements under realistic assumptions.

Nevertheless, there is more to do. Our theorems are sharpest in the high-privacy regime (small $\epsilon$'s), which may be too conservative to provide sufficient utility for some applications. While significantly relaxed from previous work, our assumptions will still not hold in all real-world systems. Thus, we hope this work encourages further collaboration between distributed systems and DP theory researchers in establishing protocols that address the full range of possible systems constraints as well as improving the full breadth of the privacy vs. utility Pareto frontier.

%% file: acks.tex
\section*{Acknowledgements}
\label{sec:acks}
The authors would like to thank Vitaly Feldman for suggesting the idea of privacy accounting in DP-SGD via shuffling, and for help in identifying and fixing a mistake in the way a previous version of this paper handled $(\eps_0, \delta_0)$-DP local randomizers.

%% file: app_proofs.tex
\section{Omitted Results and Proofs}
\label{sec:proofs}
\begin{lem}
Let $\calA_{ldp}: \calD \rightarrow \calS$ be an $\eps_0$-DP local randomizer. For $D = (d_1,\ldots, d_m) \in \calD^m, q \in (0,1)$, and $k \in [m]$, define $\BiasedSampling_q(D, k)$ to return $d_k$ with probability $q$, and a sample from an arbitrary distribution over $D \setminus \{d_k\}$ with probability $1-q$.
For any $k \in [m]$ and any set of outcomes $S \subseteq \calS$, we have
$$\frac{\Pr{[\calA_{ldp}(d_k) \in S]}}{\Pr{\sqbracket{\calA_{ldp}\bracket{\BiasedSampling_q(D, k)} \in S}}} 
\leq \frac{e^{\eps_0}}{1 + q (e^{\eps_0}-1)}. $$ 
\label{lem:pr}
\end{lem}
\begin{proof}
Fix a set of outcomes $S \subseteq \calS$. By $\eps_0$-LDP of $\calA_{ldp}$, for any $d, d' \in \calD$, we get
\begin{equation}
    \frac{\Pr[\calA_{ldp}(d) \in S]}{\Pr[\calA_{ldp}(d') \in S]} \leq e^{\eps_0} \label{eqn:eps_0b}
\end{equation} 
Now, for dataset $D = (d_1,\ldots, d_m) \in \calD^n$ and $k \in [m]$, we have:
\begin{align*}
  \frac{\Pr{[\calA_{ldp}(d_k) \in S]}}{\Pr{[\calA_{ldp}(\BiasedSampling_q(D, k)) \in S]}} & =  \frac{\Pr{[\calA_{ldp}(d_k) \in S]}}{\sum\limits_{j = 1}^m\Pr[\calA_{ldp}(d') \in S]\Pr[d' = d_j]}  \\
  & = \frac{1}{\sum\limits_{j = 1}^m\frac{\Pr[\calA_{ldp}(d') \in S]}{\Pr{[\calA_{ldp}(d_k) \in S]}}\Pr[d' = d_j]} \\
  & = \frac{1}{q + \sum\limits_{j \neq k}\frac{\Pr[\calA_{ldp}(d') \in S]}{ \Pr{[\calA_{ldp}(d_k) \in S]}}\Pr[d' = d_j]} \\
  & \leq \frac{1}{q + e^{-\eps_0} \sum\limits_{j \neq k} \Pr[d' = d_j]} \\
  & = \frac{1}{q + (1 - q)e^{-\eps_0}} = \frac{e^{\eps_0}}{1 + q(e^{\eps_0} - 1)}
\end{align*}
where the third equality follows as $\Pr[d = d_k] = q$, and the first inequality follows using inequality~\ref{eqn:eps_0b}, and the fourth equality follows as $\sum\limits_{j \neq k} \Pr[d = d_j] = 1 - q$.
\end{proof}

\begin{lem}
Let $\calA^{(1)}, \ldots, \calA^{(k)}$ be mechanisms of the form $\calA^{(i)}: \mathcal{S}^{(1)} \times \cdots \times \mathcal{S}^{(i - 1)} \times \calD \rightarrow \mathcal{S}^{(i)}$.
Suppose there exist constants $a > 0$ and $b \in (0,1)$ such that each $\calA^{(i)}$ is $\eps_i$-DP with $\eps_i \leq \log\left(1 + \frac{a}{k - b (i-1)}\right)$.
Then, for any $\delta \in (0, 1)$, the $k$-fold adaptive composition of $\calA^{(1)}, \ldots, \calA^{(k)}$ is $\left(\eps , \delta \right)$-DP with
$\eps = \frac{a^{2}}{2k(1 - b)} + \sqrt{ \frac{2a^{2} \log{(1/\delta)}}{k(1 - b)}}$.
\label{lem:het}
\end{lem}
\begin{proof}
We start by applying the heterogeneous advanced composition for DP \cite{KOV17} for the sequence of mechanisms  $\calA_1, \ldots, \calA_k$ to get $\left(\eps, \delta\right)$-DP for the composition, where
\begin{equation}
    \eps = \sum\limits_{i\in[k]} \frac{(e^{\eps_i} - 1) \eps_i}{e^{\eps_i} + 1} + \sqrt{2 \log{\frac{1}{\delta}} \sum\limits_{i\in[k]}\eps_i^2} \label{eqn:heta}
\end{equation}
Let us start by bounding the second term in equation~\ref{eqn:heta}. First, observe that:
\begin{align}
    \sum\limits_{i\in[k]}\eps_i^2 & = \sum\limits_{i\in[k]} \left(\log\left(1 + \frac{a}{k - b(i-1)}\right)\right)^2 \leq  \sum\limits_{i\in[k]} \frac{a^{2}}{(k - b(i-1))^2} \label{eqn:sq_epsa}
\end{align}
where the first inequality follows from $\log(1 + x) \leq x$.
    
Now, we have:
\begin{align*}
    \sum\limits_{i\in[k]} \frac{a^{2}}{(k - b(i-1))^2} 
    & =  \sum_{i=0}^{k-1} \frac{a^{2}}{(k - i b)^2}  \leq a^{2}  \int_{0}^{k} \frac{1}{(k - xb)^2} dx \\
    & = a^{2}\left( \frac{1}{kb - b^2k} - \frac{1}{kb} \right) = \frac{a^{2}}{kb} \left(\frac{1}{1 - b} - 1 \right) \\
    & = \frac{a^{2}}{k(1 - b)}  \numberthis \label{eqn:suma} 
\end{align*}
where the second equality follows as we have $\int \frac{1}{(c-dx)^2}dx = \frac{1}{cd - d^2x}$.

Next, we bound the first term in equation~\ref{eqn:heta} as follows:
\begin{align*}
\sum\limits_{i\in[k]} \frac{(e^{\eps_i} - 1) \eps_i}{e^{\eps_i} + 1} & = \sum\limits_{i\in[k]} \frac{\left(\frac{a}{k - b(i-1)}\right) \left(\log\left(1 + \frac{a}{k - b(i-1)}\right)\right)}{2 + \frac{a}{k - b(i-1)}}  \leq \sum\limits_{i\in[k]} \frac{\left(\frac{a}{k - b(i-1)}\right)^2}{2 + \frac{a}{k - b(i-1)}} \\
&\leq \sum\limits_{i\in[k]} \frac{a^2}{2 \left(k - b(i-1)\right)^2}
\leq \frac{a^{2}}{2k(1 - b)} \numberthis \label{eqn:sum2a} 
\end{align*}
where the first inequality follows from $\log(1 + x) \leq x$, and the last inequality follows from inequality~\ref{eqn:suma}.

Using inequalities \ref{eqn:sq_epsa}, \ref{eqn:suma} and \ref{eqn:sum2a} in equation~\ref{eqn:heta}, we get that the  $k$-fold adaptive composition of $\calA_1, \ldots, \calA_k$ satisfies $\left(\eps, \delta\right)$-DP, for 
$\eps = \frac{a^{2}}{2k(1 - b)} + \sqrt{ \frac{2a^{2} \log{(1/\delta)}}{k(1 - b)}}$.
\end{proof}

\begin{lem}\label{lem:cheu}
Suppose $\calA : \calD \rightarrow \calS$ is an $(\eps_0, \delta_0)$-DP local randomizer with $\delta_0 \leq \frac{(1 - e^{-\eps_0}) \delta_1}{4 e^{\eps_0} \left(2 + \frac{\ln(2/\delta_1)}{\ln(1/(1-e^{-5\eps_0}))}\right)}$.
Then there exists an $8 \eps_0$-DP local randomizer $\tilde{\calA} : \calD \rightarrow \calS$ such that for any $d \in \calD$ we have $TV(\calA(d), \tilde{\calA}(d)) \leq \delta_1$.
\end{lem}
\begin{proof}
The proof is a direct application of results by Cheu et al.~\cite{dpmixnets}.
First we recall that from \cite[Claims D.2 and D.5]{dpmixnets} (applied with $n = 1$ in their notation) it follows that given $\calA$ there exist randomizers $\tilde{\calA}_{k,T}$ which are $8 \eps_0$-DP and satisfy
$$
TV(\calA(d), \tilde{\calA}_{k,T}(d)) \leq \left(1 - \frac{k e^{-2 \eps_0}}{2}\right)^T + (T+2) \frac{2 \delta_0 e^{\eps_0}}{1-e^{-\eps_0}}
$$
for any $k \in (0, 2 e^{-2\eps_0})$ and $T \in \mathbb{N}$ as long as $\delta_0 < \frac{1 - e^{-\eps_0}}{4 e^{\eps_0}}$.
The result follows from taking $k = 2 e^{-3 \eps_0}$, $T = \ln(2/\delta_1) / \ln(1/(1-e^{5\eps_0}))$ and noting these choices imply the desired condition on the total variation distance under our assumption on $\delta_0$.
\end{proof}

\begin{proof}[Proof of Corollary~\ref{cor:res}]
Setting $p_0=\frac{m}{n}$ in $\calA_{fix}$, we get from Theorem~\ref{thm:ampl_dist+} that $\beta \in (0,1)$, algorithm $\calA_{fix}$ satisfies $(\eps_1, \beta)$-DP for 
\begin{align*}
    \eps_1 & = \frac{(e^{\eps_0} - 1)\sqrt{2m e^{\eps_0} \log{(1/\beta)}}}{n} + \frac{m e^{\eps_0} (e^{\eps_0} - 1)^2}{2n^2} \\
    & \leq \frac{2(e^{\eps_0} - 1)\sqrt{2m e^{\eps_0} \log{(1/\beta)}}}{n} \numberthis \label{eq: eps_1}
\end{align*}
where the inequality follows since $n \geq (e^{\eps_0} - 1)\sqrt{m e^{\eps_0}}$.

Now, using inequality~\ref{eq: eps_1} and applying advanced composition to $\frac{n}{m}$ repetitions of $\calA_{fix}$, we get $\bracket{\eps, \frac{n\beta}{m} + \delta}$-DP, for
\begin{align}
    \eps &\leq  \eps_1\sqrt{\frac{2n}{m}\log\bracket{1/\delta}} + \frac{n}{m}\eps_1 \bracket{e^{\eps_1} - 1} \label{eq: adv}
\end{align}
Since $\eps_0 \leq \frac{2 \log{\bracket{n/8\sqrt{m}}}}{3}$, we have that $\eps_1 \leq \frac{1}{2}$, and thus, $\bracket{e^{\eps_1} - 1} \leq \frac{3\eps_1}{2}$.
Therefore, we get from inequality~\ref{eq: adv} that
\begin{align*}
    \eps & \leq \eps_1\sqrt{\frac{2n}{m}\log\bracket{1/\beta}} + \frac{3n}{2m}\eps_1^2 \\
    & \leq 4(e^{\eps_0} - 1)\sqrt{\frac{ e^{\eps_0} \log{(1/\beta)}\log\bracket{1/\delta}}{n}} + \frac{12(e^{\eps_0} - 1)^2 e^{\eps_0} \log{(1/\beta)}}{n} \\
    & = \widetilde{O}\bracket{\frac{e^{1.5\eps_0}}{\sqrt{n}}}
\end{align*}
where the equality holds since $n \geq (e^{\eps_0} - 1)^2e^{\eps_0}\sqrt{m}\log\bracket{1/\beta}$, and $\tilde{O}(\cdot)$ hides polylog factors in $1/ \beta$ and $1/ \delta$.
\end{proof}

\begin{proof}[Proof of Proposition~\ref{prop:dummy_dist+}]
 In Algorithm~\ref{alg:a_dist_new}, for $i \in [m]$, we have  $$S_i = \{j : \text{User}(j) \text{ checks-in for index } i \}$$ For $i \in [m]$, define an indicator random variable $E_i$ that indicates if $S_i$ is empty. 
 Note that the server performs a dummy gradient update for instance $i \in [n]$ if and only if $S_{i}$ is empty (or, in other words, $E_i = 1$). 
 Next, for $j \in [n]$, let $I_j$ denote the index that user $j$ in Algorithm $\calA_{fix}$ performs her $(R_j, p_j)$-check-in into, where $R_j = [m]$ and $p_j = p_0$. Thus, for index $i \in [m]$, we have 
\begin{align*}
    \Pr{\left[E_i = 1\right]} & =
    \Pr{\left[ \bigcap\limits_{j \in [n]} \left( \bracket{\text{User } j \text{ abstains}} \bigcup \bracket{\text{User } j \text{ participates} \land I_j \neq i}   \right)  \right]}\\
    & = \prod\limits_{j \in [n]} \bracket{(1 - p_0) + \Pr{\left[ I_j \neq i \right]} \cdot p_0}  = \left((1 - p_0) +  \left(1 - \frac{1}{m}\right)\cdot p_0   \right)^n \\ 
    & =  \left(1 - \frac{p_0}{m}\right)^n 
\end{align*}
where the second equality follows since the check-ins for each user are independent of the others, and each user abstains from participating w.p. $(1 - p_0)$.

Thus, for the expected number of dummy gradient updates, we have:
\begin{align}
\mathbb{E}(E_{1:m}) & = \sum\limits_{i \in [m]} \Pr{\left[E_i = 1\right]} =  m  \left(1 - \frac{p_0}{m}\right)^n \label{eqn:p_z}
\end{align}

If $p_0 = \frac{cm}{n}$ for $c > 0$, from equation~\ref{eqn:p_z} we get 
\begin{align*}
\mathbb{E}(E_{1:m}) & = m  \left(1 - \frac{c}{n}\right)^n \leq \frac{m}{e^c}
\end{align*}
where the inequality follows as $\left(1 - \frac{a}{b}\right)^b \leq e^{-a}$ for $b > 1, |a| \leq b$.
\end{proof}

\begin{proof}[Proof of Theorem~\ref{thm:util_new}]
To be able to directly apply \cite[Theorem 2]{shamir2013stochastic}, our technique $\calA_{fix}$ needs to satisfy two conditions: i) each model update should be an unbiased estimate of the gradient, and ii) a bound on the expected $L_2$-norm of the gradient.
Notice that in $\calA_{fix}$, every client $j \in [n]$ performs a $([m],p_0)$-check-in. 
This is analogous to a bins-and-balls setting where $n$ balls are thrown, each with probability $p_0$, into $m$ bins. Thus, for each update step $i \in [m]$, the number of clients checking-in for this step (i.e., $|S_i|$ in the notation of Algorithm~\ref{alg:a_dist_new}) can be approximated by an independent Poisson random variable $Y_i$ with mean $np_0 / m$, using Poisson approximation \cite{MU}, as follows: 
 $$ \Pr[|S_i| = 0] \leq 2\Pr[Y_i = 0] = 2e^{-np_0 / m} \coloneqq p'$$ 
 
 Now, we know that there exists a probability $p_b \leq p'$ with which the gradient update $g_i$ is $0^p$. Thus, to make the gradient update unbiased, each participating user can multiply their update by $\frac{1}{1 - p_b} \leq \frac{1}{1 - p'} = \frac{1}{1 - 2e^{-np_0 / m}}$.  Consequently, the Lipschitz-constant of the loss $\ell$, and the variance of the noise added to the update, increases by a factor of at most $\frac{1}{\left(1 - 2e^{-np_0 / m}\right)^2}$. Thus, we get $\mathbb{E}\sqbracket{||\widetilde{g}_i||^2} \leq \frac{p\sigma^2 + L^2}{1 - 2e^{-np_0 / m}}$.
 With this, our technique will satisfy both the conditions required to apply the result in \cite{shamir2013stochastic}  for learning rate $\eta_i = \frac{c}{\sqrt{i}}$ as follows:
 $$\mathbb{E}_{D,\theta_m}\left[\mathscr{L}(\mathscr{D}; \theta_m) \right] - \mathscr{L}(\mathscr{D}; \theta^*) \leq \left(\frac{R^2}{c} + \frac{c(p\sigma^2 + L^2)}{1 - 2e^{-np_0 / m}} \right) \left(\frac{2 + \log(m)}{\sqrt{m}}\right) $$
 Optimizing the learning rate to be $\eta_i = \frac{R \left(1 - 2e^{-n p_0 / m}\right)}{\sqrt{\bracket{p\sigma^2 + L^2}i}}$ gives the statement of the theorem.
\end{proof}

\begin{proof}[Proof of Theorem~\ref{thm:util_new2}]
We prove the first bound on the line of the proof of Theorem~\ref{thm:util_new}.
Since $\calA_{avg}$ skips an update for time step $i \in [m]$ if no client checks-in at step $i$, and otherwise makes an update of the average of the noisy gradients received by checked-in clients, each update of the algorithm is unbiased.
Now, notice that in $\calA_{avg}$, every client $j \in [n]$ checks into $[m]$ u.a.r. Thus, each update step $i \in [m]$ will have $n/m$ checked-in clients in expectation. As a result, for an averaged update $\widetilde{h}_{i} = \frac{\widetilde{g}_i}{|S_i|}$, we get $\mathbb{E}\sqbracket{\norm{\widetilde{h}_i}^2} \leq \frac{mp\sigma^2}{n} + L^2 $.
With this, our technique will satisfy both the conditions required to apply \cite[Theorem 2]{shamir2013stochastic}  for learning rate $\eta_i = \frac{c}{\sqrt{i}}$, giving:
$$\mathbb{E}_{D,\theta_m}\left[\mathscr{L}(\mathscr{D}; \theta_m) \right] - \mathscr{L}(\mathscr{D}; \theta^*) \leq \left(\frac{R^2}{c} + c \bracket{\frac{mp\sigma^2}{n} + L^2}\right) \left(\frac{2 + \log(m)}{\sqrt{m}}\right) $$
Optimizing the learning rate to be $\eta_i = \frac{R\sqrt{n}}{\sqrt{\bracket{mp \sigma^2 + nL^2}i}}$ gives the statement of the theorem.

When in addition the loss is $\beta$-smooth we can obtain an improved bound on the expected risk -- in this case, for the average parameter vector $\frac{1}{m} \sum_i \theta_i$ -- by applying \cite[Theorem 6.3]{bubeck2015convex}.
Let $h_i = \nabla_{\theta} \mathscr{L}(\mathscr{D}; \theta_i)$ be the true gradient on the population loss at each iteration.
The cited result says that after $m$ iterations with learning rate $\eta_i = \frac{1}{\beta + \frac{\kappa \sqrt{t}}{\sqrt{2} R}}$ with $\kappa^2 \geq \mathbb{E}[\norm{h_i - \tilde{h}_i}^2]$ we get
$$\mathbb{E}_{D,\theta_1, \ldots, \theta_m}\left[\mathscr{L}\left(\mathscr{D}; \frac{1}{m} \sum_{i = 1}^m \theta_i\right) \right] - \mathscr{L}(\mathscr{D}; \theta^*)
\leq
R \kappa \sqrt{\frac{2}{m}} + \frac{\beta R^2}{m}
$$
The result now follows from observing that
\begin{align*}
    \mathbb{E}[\norm{h_i - \tilde{h}_i}^2] &\leq
    \mathbb{E}_{S \sim \mathrm{Bin}(n, 1/m)}\left[\frac{1}{S} (L^2 + p \sigma^2) \middle| S > 0 \right]
    = O\left(\frac{m}{n}(L^2 + p \sigma^2)\right)
\end{align*}
\end{proof}

\begin{proof}[Proof of Proposition~\ref{prop:dummy_dist}]
In Algorithm $\calA_{sldw}$, for $i \in [n - m + 1]$, we have $$S_i = \{j : \text{User}(j) \text{ checks-in for index } i \}$$ For $i \in [n]$, define an indicator random variable $E_i$ that indicates if $S_i$ is empty. 
Note that the server performs a dummy gradient update for instance $i \in [n]$ if and only if $S_{i}$ is empty (or, in other words, $E_i = 1$). 
Next, for $j \in [m]$, let $I_j$ denote the index that user $j$ in Algorithm $\calA_{fix}$ performs her $R_j$-check-in into, where $R_j = \{j, \ldots, j+m-1\}$.  Thus, for index $i \in \{m, \ldots, n - m + 1\}$, we have 
\begin{align}
    \Pr{\left[E_i = 1\right]} & = \Pr{\left[ \bigcap\limits_{j \in [ i - m + 1, i]} I_j \neq i \right]}   = \prod\limits_{j \in [i - m + 1, i]} \Pr{\left[ I_j \neq i \right]}  = \left(1 - \frac{1}{m} \right)^m \leq \frac{1}{e} \label{eqn:dum1}
\end{align}
where the second equality follows since the check-ins for each user are independent of the others, and the inequality follows as $\left(1 - \frac{a}{b}\right)^b \leq e^{-a}$ for $b > 1, |a| \leq b$.

Thus, for the expected number of dummy gradient updates, we have:
\begin{align*}
\mathbb{E}(E_{1:n}) & = \sum\limits_{i \in \{m, \ldots, n - m + 1\}} \Pr{\left[E_i = 1\right]} \leq \frac{n - m + 1}{e}
\end{align*}
where the inequality follows from inequality~\ref{eqn:dum1}.
\end{proof}

\subsection{Proof of Theorems \ref{thm:ampl_dist+} and \ref{thm:ampl_dist}}
\label{sec:proof_ampl_dist}

We will first prove the privacy guarantee of $\calA_{fix}$ (Algorithm~\ref{alg:a_dist_new}) 
by reducing it to algorithm $\calA_{rep}$ (Algorithm~\ref{alg:a_rep_new}) that starts by swapping the first element in the dataset by a given replacement element, randomly chooses a position in the dataset to get replaced by the original first element with a given probability, and then carries out DP-SGD with the local randomizer. 
W.l.o.g., for simplicity we will define $\calA_{rep}$ to update the model for 1-sized minibatches (i.e., update at every time step). It is easy to extend to $b$-sized minibatch updates by accumulating the gradient updates for every $b$ steps and then updating the model. 

For the proofs that follow, it will be convenient to define additional notation for denoting distance between distributions. Given 2 distributions $\mu$ and $\mu'$, we denote them as $\mu \approxeq_{(\eps, \delta)}\mu'$ if they are $(\eps, \delta)$-DP close, i.e., if for all measurable outcomes $S$, we have 
$$e^{-\eps} \bracket{\mu'(S) - \delta} \leq \mu(S) \leq e^{\eps}\mu'(S) + \delta $$

\begin{algorithm}[ht]
	\caption{$\calA_{rep}$: DP-SGD with One Random Replacement}
	\begin{algorithmic}[1]
	    \REQUIRE Dataset $D=d_{1:m}$, local randomizer $\calA_{ldp}$. \\
	    \textbf{Parameters: } Initial model $\theta_1 \in \mathbb{R}^p$, weights $w_{1:m}$ where $w_i \in [0,w_{max}]$ for $i \in [m]$, replacement element $d_r$
	    \STATE Sample $I \xleftarrow{u.a.r.} [m]$
	    \STATE Let $G \leftarrow (d_r, d_{2:m})$
	    \STATE Let $\sigma_I(D) \leftarrow (G_{1:I-1}, z_I, G_{I+1:m})$, where   $z_I = \begin{cases}
             d_1  & \text{with probability } w_I \\
             G[I]  & \text{otherwise}
       \end{cases}$
       \FOR{$i \in [m]$}
            \STATE $\tilde{g}_i \leftarrow \calA_{ldp}(\theta_{i};\sigma_I(D)[i])$
            \STATE $\theta_{i+1} \gets \theta_i - \eta \tilde{g}_i$
            \STATE Output $\theta_{i + 1}$
       \ENDFOR
	\end{algorithmic}
	\label{alg:a_rep_new}
\end{algorithm}

\begin{thm}[Amplification via random replacement]
Suppose $\calA_{ldp}$ is an $\eps_0$-DP local randomizer. 
Let $\calA_{rep}: \calD^{m} \rightarrow \Theta^{m}$ be the protocol from Algorithm~\ref{alg:a_rep_new}.
For any $\delta \in (0,1)$, algorithm $\calA_{rep}$ is $\left(\eps, \delta\right)$-DP at index 1 in the central model, where $\eps = \frac{w_{max}^2 e^{\eps_0} (e^{\eps_0} - 1)^2}{2m} + w_{max}(e^{\eps_0} - 1)\sqrt{\frac{2 e^{\eps_0} \log{(1/\delta)}}{m}}$. In particular, for $\eps_0 \leq 1$ and $\delta \leq 1/100$, we get $\eps \leq 7w_{max}\eps_0\sqrt{\frac{\log(1/\delta)}{m}}$. Here, initial model $\theta_1 \in \mathbb{R}^p$, weights $w_{max}\in [0, 1]$, $w_i \in [0,w_{max}]$ for every $i \in [m]$, and replacement element $d_r \in [0,1]$ are parameters to $\calA_{rep}$.
Furthermore, if $\calA_{ldp}$ is an $(\eps_0,\delta_0)$-DP local randomizer with $\delta_0 \leq \frac{(1 - e^{-\eps_0}) \delta_1}{4 e^{\eps_0} \left(2 + \frac{\ln(2/\delta_1)}{\ln(1/(1-e^{-5\eps_0}))}\right)}$, then algorithm $\calA_{rep}$ is $\left(\eps', \delta'\right)$-DP at index 1 in the central model, where $\eps' = \frac{w_{max}^2 e^{8\eps_0} (e^{8\eps_0} - 1)^2}{2m} + w_{max}(e^{8\eps_0} - 1)\sqrt{\frac{2 e^{8\eps_0} \log{(1 / \delta)}}{m}}$ and $\delta' = \delta + m (e^{\eps'}+1) \delta_1$.
\label{thm:ampl_rep_new}
\end{thm}

\begin{proof}
We start by proving the privacy guarantee of $\calA_{rep}$ for the case where the local randomizer $\calA_{ldp}$ is $\eps_0$-DP, i.e., for the case where $\delta_0 = 0$.
Let us denote the output sequence of $\calA_{rep}$ by $Z_2, Z_3, \ldots , Z_{m+1}$. 
Note that $Z_{2:m+1}$ can be seen as the output of a sequence of $m$ algorithms with conditionally independent randomness: $\calB^{(i)}$ for $i \in [m]$ as follows.
On input $\theta_{2:i}$ and $D$, $\calB^{(i)}$ outputs a random sample from the distribution of $Z_{i+1}|Z_{2:i}=\theta_{2:i}$. 
The outputs of $\calB^{(1)}, \ldots, \calB^{(i-1)}$ are given as input to $\calB^{(i)}$. 
Therefore, in order to upper bound the privacy parameters of $\calA_{rep}$, we analyze the privacy parameters of $\calB^{(1)}, \ldots, \calB^{(m)}$ and apply the heterogeneous advanced composition for DP \cite{KOV17}.

Next, observe that conditioned on the value of $I$, $Z_{i+1}$ is the output of $\calA^{(i)}_{ldp}(\theta_i;d)$ with its internal randomness independent of $Z_{2:i}$. In particular, for $i \geq 2$, one can implement $\calB^{(i)}$ as follows. First, sample an index $T$ from the distribution of $I|Z_{2:i}=\theta_{2:i}$. 
Assign $\tilde{g}_i = \calA_{ldp}(\theta_i;d_1)$ w.p. $w_i$ if $T=i$, otherwise let $\tilde{g}_i = \calA_{ldp}(\theta_i;d_i)$. For $\calB^{(1)}$, we first sample $T$ u.a.r. from $[m]$, and let $\tilde{g}_1 = \calA_{ldp}(\theta_1; d_1)$ w.p. $w_1$ if $T=1$, otherwise let $\tilde{g}_1 = \calA_{ldp}(\theta_1; d_r)$. For each $i \in [m]$, algorithm $\calB^{(i)}$ outputs $\theta_{i + 1} = \theta_i - \eta \tilde{g}_i$.

We now prove that for each $i\in[m]$, $\calB^{(i)}$ is $\left(\log\left(1 + \frac{w_{max}e^{\eps_0}(e^{\eps_0}-1)}{i-1 + e^{\eps_0}(m - i + 1)}\right),0\right)$-DP at index 1. 
Let $D=d_{1:m}$ and $D'=(d'_1, d_{2:m})$ be 2 datasets differing in the first element. 
Let $\theta_{2:i}$ denote the input to $\calB^{(i)}$. Let $\mu$ be the probability distribution of $\calB^{(i)}(\theta_{2:i};D)$. 
Let $\mu_1$ be the distribution of $\calB^{(i)}(\theta_{2:i};D)$ conditioned on  $\tilde{g}_i = \calA_{ldp}(\theta_{2:i};d_1)$, and $\mu_0$ be the distribution of $\calB^{(i)}(\theta_{2:i};D)$ conditioned on $\tilde{g}_i =\calA_{ldp}(\theta_{2:i};d_r)$ for $i = 1$, and $\tilde{g}_i = \calA_{ldp}(\theta_{2:i};d_i)$ for $i \geq 2$.  Also, denote by $\mu', \mu_0'$, and $\mu_1'$ the corresponding quantities when $\calB^{(i)}$ is run on $D'$. 
Let $q_i$ be the probability that $T=i$ (sampled from $I|Z_{2:i}=\theta_{2:i}$). By definition, $\mu = (1-q_iw_i)\mu_0 + q_iw_i \mu_1$, as $\calB^{(i)}(s_{1:i-1};D)$ generates output using $\calA_{ldp}(\theta_{2:i};d_1)$ w.p. $w_i$ if $T = i$.  Similarly, $\mu' = (1-q_i' w_i)\mu_0' + q_i' w_i\mu_1'$ when the input dataset is $D'$.

For $i\in[m]$, we observe that $\mu_0=\mu_0'$, since in both cases the output is generated by $\calA_{ldp}(\theta_1, d_r)$ for $i = 1$, and $\calA_{ldp}(\theta_{2:i};d_i)$ for $i \geq 2$.
W.l.o.g. assume that $q_i \geq q_i'$.
Thus, we can shift $(q_i - q_i') w_i$ mass from the first component of the mixture in $\mu'$ to the second component to obtain
$$\mu' = (1-q_i w_i) \mu_0 + q_i w_i \left(\frac{q_i'}{q_i} \mu_1' + \left(1 - \frac{q_i'}{q_i}\right)\mu_0\right)
= (1-q_i w_i ) \mu_0 + q_i w_i \mu_1''$$
This shows that $\mu$ and $\mu'$ are overlapping mixtures \cite{BBG18}.
Now, $\eps_0$-LDP of $\calA_{ldp}$ implies $\mu_0 \approxeq_{(\eps_0,0)} \mu_1$ and $\mu_0' \approxeq_{(\eps_0,0)} \mu_1'$. Moreover, $\eps_0$-LDP of $\calA_{ldp}$ also implies $\mu_1 \approxeq_{(\eps_0,0)} \mu_1'$, so by the joint convexity of the relation $\approxeq_{(\eps_0,0)}$ we also have $\mu_1 \approxeq_{(\eps_0,0)} \mu_1''$.
Thus, we can apply Advanced Joint Convexity of overlapping mixtures (Theorem 2 in \cite{BBG18}) to get that
\begin{equation}
    \mu \approxeq_{(\log\left(1 + q_i w_i(e^{\eps_0} - 1)\right),0)} \mu' \label{eqn:mu}
\end{equation}

We now claim that $q_i\leq \frac{e^{\eps_0}}{i-1 + e^{\eps_0}(m - i + 1)}$. Observe that for each $D^* \in \{ D, D'\}$, conditioning on $T=i$ reduces $\calA_{rep}$ to running $\calA_{ldp}$ on $\sigma_i(D^*)$. Note that for $j < i$, we have that $\sigma_i(D^*)[1:i-1]$ differs from $\sigma_j(D^*)[1:i-1]$ in at most 1 position, and for $j > i$, we have $\sigma_i(D^*)[1:i-1] = \sigma_j(D^*)[1:i-1]$. Since $\Pr[j \geq i] = \frac{m - i + 1}{m}$, by setting $q = \frac{m - i + 1}{m}$ in Lemma~\ref{lem:pr}, we get that
\begin{equation}
    \frac{\Pr[Z_{2:i}=\theta_{2:i} | T=i]}{\Pr[Z_{2:i}=\theta_{2:i}]} \leq  \frac{e^{\eps_0}}{1 + \frac{(m - i + 1)}{m}(e^{\eps_0} - 1)} = \frac{me^{\eps_0}}{i-1 + e^{\eps_0}(m - i + 1)}  \label{eqn:zi}
\end{equation}
This immediately implies our claim, since we have
\begin{align*}
    q_i & = \Pr[T=i | Z_{2:i}=\theta_{2:i}] = \frac{\Pr[Z_{2:i}=\theta_{2:i} | T=i] \cdot \Pr[t=i]}{\Pr[Z_{2:i}=\theta_{2:i}]}  \\
    & \leq \frac{e^{\eps_0}}{i-1 + e^{\eps_0}(m - i + 1)}
\end{align*}
where the inequality follows from inequality~\ref{eqn:zi}, and as $\Pr[T=i] = \frac{1}{m}$.

Substituting the value of $q_i$ in equation~\ref{eqn:mu}, and using the fact that $w_i \leq w_{max}$, we get that for each $i\in[m]$, algorithm $\calB^{(i)}$ is $\left(\eps_i,0\right)$-DP at index 1, where $\eps_i = \log\left(1 + \frac{w_{max}e^{\eps_0}(e^{\eps_0}-1)}{i-1 + e^{\eps_0}(m - i + 1)}\right) $.  This can alternatively be written as $\eps_i = \log\left(1 + \frac{w_{max}(e^{\eps_0}-1)}{m - (i-1) \frac{e^{\eps_0} - 1}{e^{\eps_0}}}\right)$, and using Lemma~\ref{lem:het} for the sequence of mechanisms  $\calB^{(1)}, \ldots, \calB^{(m)}$ by setting $a = w_{max}(e^{\eps_0}-1)$, $b=\frac{e^{\eps_0} - 1}{e^{\eps_0}}$, and $k = m$, we get that
algorithm $\calA_{rep}$ satisfies $\left(\eps, \delta\right)$-DP at index 1, for 
$\eps = \frac{w_{max}^2e^{\eps_0} (e^{\eps_0} - 1)^2}{2m} + w_{max}(e^{\eps_0} - 1)\sqrt{\frac{2 e^{\eps_0} \log{(1/\delta)}}{m}}$.

Now, for the above bound, if $\eps_0 \leq 1$ and $\delta \leq 1/4$, we get that
\begin{align*}
    \eps & = \frac{w_{max}^2e^{\eps_0} (e^{\eps_0} - 1)^2}{2m} + w_{max}(e^{\eps_0} - 1)\sqrt{\frac{2 e^{\eps_0} \log{(1 / \delta)}}{m}} \\
    & = \frac{w_{max}e^{0.5\eps_0} (e^{\eps_0} - 1)}{\sqrt{m}} \bracket{\frac{w_{max}e^{0.5\eps_0} (e^{\eps_0} - 1)}{2\sqrt{m}} + \sqrt{2 \log{(1 / \delta)}}}  \\
    & \leq \frac{3w_{max}\eps_0}{\sqrt{m}} \bracket{\frac{3w_{max}\eps_0}{2\sqrt{m}} + \sqrt{2 \log{(1 / \delta)}}} \\
    & \leq \frac{3w_{max}\eps_0}{\sqrt{m}} \bracket{\bracket{\sqrt{2} + \sqrt{1/2}} \sqrt{ \log{(1 / \delta)}}} \leq 7w_{max}\eps_0\sqrt{\frac{\log{(1 / \delta)}}{m}} 
\end{align*}
where the first inequality follows since $e^{0.5\eps_0}(e^{\eps_0} - 1) \leq 3 \eps_0$ for $\eps_0 \leq 1$, and the second inequality follows since $\frac{3w_{max}\eps_0}{2\sqrt{m}} \leq \sqrt{\frac{\log{\frac{1 }{\delta}}}{2}}$ for $\delta \leq 1/100$.

Now, we prove the privacy guarantee of $\calA_{rep}$ for the more general case where for each $i \in [m]$, the local randomizer $\calA_{ldp}$ is $(\eps_0, \delta_0)$-DP. To upper bound the privacy parameters of $\calA_{rep}$, we modify the local randomizer to satisfy pure DP, apply the previous analysis, and then account for the difference between the protocols with original and modified randomizers using the total variation distance.

Since $\calA_{ldp}$ is $(\eps_0, \delta_0)$-DP with $\delta_0 \leq \frac{(1 - e^{-\eps_0}) \delta_1}{4 e^{\eps_0} \left(2 + \frac{\ln(2/\delta_1)}{\ln(1/(1-e^{-5\eps_0}))}\right)}$, we get from Lemma~\ref{lem:cheu} that there exists a randomizer $\tilde{\calA}_{ldp}$ that is $8 \eps_0$-DP, and for any data record $d$ and parameter vector $\theta$ satisfies $TV\left(\calA_{ldp}(d; \theta), \tilde{\calA}_{ldp}(d; \theta)\right) \leq \delta_1$.
After replacing every instance of $\calA_{ldp}$ in $\calA_{rep}$ with $\tilde{\calA}_{ldp}$ to obtain $\tilde{\calA}_{rep}$, a union bound gives:
\begin{equation}
    TV\left(\calA_{rep}(D); \tilde{\calA}_{rep}(D)\right) \leq m \delta_1 \label{eqn:tv}
\end{equation}

Now, proceeding in a similar manner as in the case of $\eps_0$-DP local randomizers above to see that $\tilde{\calA}_{rep}$ using the $8 \eps_0$-DP local randomizers $\tilde{\calA}_{ldp}$ satisfies $(\eps', \delta)$-DP at index 1 with $\eps' = \frac{w_{max}^2 e^{8\eps_0} (e^{8\eps_0} - 1)^2}{2m} + w_{max}(e^{8\eps_0} - 1)\sqrt{\frac{2 e^{8\eps_0} \log{(1 / \delta)}}{m}}$.
Thus, using Proposition 3 from \cite{PFF} and inequality~\ref{eqn:tv}, we get that $\calA_{rep}$ satisfies $(\eps', \delta')$-DP at index 1 with $\delta' = \delta + m (e^{\eps'}+1) \delta_1$.
\end{proof}

Now, we are ready to prove Theorems  \ref{thm:ampl_dist+} and \ref{thm:ampl_dist}.

\begin{proof}[Proof of Theorem~\ref{thm:ampl_dist+}] 
 Let $D$ and $D'$ be 2 datasets of $n$ users that differ in a user at some index $i^* \in [n]$. Algorithm $\calA_{fix}$ can be alternatively seen as follows.  
 The server starts by initializing $F = [0^p]^m$, weights $W = [1]^m$, and for $i \in [m]$, set $S_i = \phi$. 
  For each user $j\in [n]$ s.t. $j \neq i^*$, user $j$ performs a random check-in along with some additional operations. She first samples $I_j$ u.a.r. from $[m]$, and w.p. $p_0$ does the following: she requests the server for model at index $I_j$ (and gets inserted into set $S_{I_j}$ at the server). She also updates $F[I_j] = d_j$ with probability $W[I_j]$, and sets $W[I_j] = \frac{W[I_j]}{W[I_j] + 1}$. Next, the server runs $\calA_{rep}$ on input dataset $\pi^*(D) = (d_{i^*}, F[2:m])$, with the replacement element $F[1]$, initial model $\theta_1$, and weight parameters set to $W'[1:m]$, where $W'[i] = W[i] \cdot p_0$. 
  
First, notice that in the alternative strategy above, for each of the weights $W[i], i\in [m]$, it always holds that $W[i] = \frac{1}{\left\vert S_i\right\vert}$. Thus, each weight $W[i], i\in [m]$  is updated to simulate reservoir sampling~\cite{V85} of size 1 in slot $F[i]$. In other words, updating $F[i] = d$ with probability $W[i]$ for an element $d$ is equivalent to $F[j] \xleftarrow{u.a.r.} S_i$, where $S_i$ is the set containing $d$ and all the elements previously considered for updating $S_i$.
As a result, since the first element in $\calA_{rep}$ performs a random  replacement with weights set to $W'[1:m]$ for its input dataset, it is easy to see that performing a concurrent random check-in for user $i^*$ (as in Algorithm~\ref{alg:a_dist_new}) is equivalent to performing a random replacement for her after the check-ins of all the other users. 

From our construction, we know that datasets $\pi^*(D)$ and $\pi^*(D')$, which are each of length $m$, differ only in the element with index 1. Moreover, in the alternative strategy above, note that the weights $W'[1: m]$ and the replacement element $F[1]$ input to $\calA_{rep}$ are independent of the data of user $i^*$ in the original dataset. Therefore, in the case $\delta_0 = 0$, using Theorem~\ref{thm:ampl_rep_new} and setting $w_{max} = p_0$, we get $\calA_{rep}(\pi^*(D)) \approxeq_{\eps, \delta} \calA_{rep}(\pi^*(D'))$ at index 1, for
 $\eps = \frac{p^2e^{\eps_0} (e^{\eps_0} - 1)^2}{2m} + \frac{p(e^{\eps_0} - 1)\sqrt{2 e^{\eps_0} \log{(1/\delta)}}}{m}$,
which implies $\calA_{dist}(D) \approxeq_{\eps, \delta} \calA_{dist}(D')$. Consequently, it implies  $\eps \leq 7p_0\eps_0\sqrt{\frac{\log(1/\delta)}{m}}$ for $\eps_0 \leq 1$ and $\delta \leq 1/100$.

The case $\delta_0 > 0$ follows from the same reduction using the corresponding setting of Theorem~\ref{thm:ampl_rep_new}.
\end{proof}

\begin{proof}[Proof of Theorem~\ref{thm:ampl_dist}]
 We proceed similar to the proof of Theorem~\ref{thm:ampl_dist+}.
 Let $D$ and $D'$ be 2 datasets of $n$ users that differ in a user at some index $i^* \in [n]$. Algorithm $\calA_{sldw}$ can be alternatively seen as follows.  
 The server starts by initializing $F = [0^p]^{n - m + 1}$, weights $W = [1]^{n - m + 1}$, and for $j \in \{m, \ldots, n\}$, set $S_j = \phi$. 
  For each user $j\in [n]$ s.t. $j \neq i^*$, user $j$ performs a random check-in along with some additional operations. She first samples $I_j$ u.a.r. from $\{j, \ldots,  j + m - 1\}$, requests the server for model at index $I_j$ (and gets inserted into set $S_{I_j}$ at the server). She also updates $F[I_j] = d_j$ with probability $W[I_j]$, and sets $W[I_j] = \frac{W[I_j]}{W[I_j] + 1}$. 
  
   Now, the server runs its loop until it releases $i^* - 1$ outputs. Next, the server runs $\calA_{rep}$ on input dataset $\pi^*(D) = (d_{i^*}, F[i^*+1:i^*+m])$, with weight parameters set to $W[i^*:i^*+m]$, initializing model $\theta_{i^*}$, and the replacement element $F[i^*]$. Lastly, the server releases the last $(n - (i^* + m) + 1)$ outputs of $\calA_{sldw}$ using $F[i^*+m + 1: n]$ and the local randomizer $\calA_{ldp}$.

 First, notice that in the alternative strategy above, for each of the weights $W[i], i\in [n]$, it always holds that $W[i] = \frac{1}{\left\vert S_i\right\vert}$. Thus, each weight $W[i], i\in [n]$  is updated to simulate reservoir sampling~\cite{V85} of size 1 in slot $F[i]$. In other words, updating $F[i] = d$ with probability $W[i]$ for an element $d$ is equivalent to $F[i] \xleftarrow{u.a.r.} S_i$, where $S_i$ is the set containing $z$ and all the elements previously considered for updating $S_i$.
As a result, since the first element in $\calA_{rep}$ performs a random replacement for its input dataset (which doesn't include $F_{1:i^* - 1} \bigcup F_{i^* + m + 1:n}$ in the alternative strategy above), it is easy to see that sequentially performing a random check-in for user $i^*$ (as in Algorithm~\ref{alg:a_dist_new}) is equivalent to performing a random replacement for her after the check-ins of all the other users and releasing the first $i^* - 1$ outputs of $\calA_{sldw}$. 

From our construction, we know that datasets $\pi^*(D)$ and $\pi^*(D')$, which are each of length $m$, differ only in the element with index 1. Moreover, in the alternative strategy above, note that the weights $W[i^*:i^* + m]$, initializing model $\theta_{i^*}$ and the replacement element $F[i^*]$ input to $\calA_{rep}$ are independent of the data of user $i^*$ in the original dataset. Therefore, using Theorem~\ref{thm:ampl_rep_new} and setting $w_{max} = 1$, we get $\calA_{rep}(\pi^*(D)) \approxeq_{\eps, \delta + m \delta_0} \calA_{rep}(\pi^*(D'))$ at index 1, for
 $\eps = \frac{e^{\eps_0} (e^{\eps_0} - 1)^2}{2m} + (e^{\eps_0} - 1)\sqrt{\frac{2 e^{\eps_0} \log{(1/\delta)}}{m}}$,
 which implies $\calA_{rc}(D) \approxeq_{\eps, \delta + m \delta_0} \calA_{rc}(D')$. Consequently, it implies  $\eps \leq 7\eps_0\sqrt{\frac{\log(1/\delta)}{m}}$ for $\eps_0 \leq 1$ and $\delta \leq 1/100$.
 
 The case $\delta_0 > 0$ follows from the same reduction using the corresponding setting of Theorem~\ref{thm:ampl_rep_new}.
\end{proof}

\input{protocol_avg}

\subsection{Proof of Theorem~\ref{thm:ampl_shuff}} 
\label{sec:proof_ampl_shuff}

\begin{algorithm}[ht]
	\caption{$\calA_{sl}$: Local responses with shuffling}
	\begin{algorithmic}[1]
	    \REQUIRE Dataset $D=d_{1:n}$, algorithms $\calA^{(i)}_{ldp}: \mathcal{S}^{(1)} \times \cdots \times \mathcal{S}^{(i - 1)} \times \calD \rightarrow \mathcal{S}^{(i )}$ for $i \in [n]$. \\
	    \STATE Let $\pi$ be a uniformly random permutation of $[n]$
	    \FOR{$i \in [n]$}
       \STATE $s_i \leftarrow \calA^{(i)}_{ldp}(s_{1:i-1}; d_{\pi(i)})$%
       \ENDFOR
       \STATE \textbf{return} sequence $s_{1:n}$
	\end{algorithmic}
	\label{alg:a_shuff}
\end{algorithm}
We will prove the privacy guarantee of $\calA_{sl}$ (Algorithm~\ref{alg:a_shuff}) in a similar manner as in the proof of Theorem 7 in \cite{soda-shuffling}: by reducing $\calA_{sl}$ to $\calA_{swap}$ that starts by swapping the first element with a u.a.r. sample in the dataset, and then applies the local randomizers (Algorithm~\ref{alg:a_swap}).
They key difference between our proof and the one in \cite{soda-shuffling} is that we provide tighter, position-dependent privacy guarantees for each of the outputs of $\calA_{swap}$, and then use an \emph{heterogeneous} adaptive composition theorem from \cite{KOV17} to compute the final privacy parameters.

\begin{algorithm}[ht]
	\caption{$\calA_{swap}$: Local responses with one swap}
	\begin{algorithmic}[1]
	    \REQUIRE Dataset $D=d_{1:n}$, algorithms $\calA^{(i)}_{ldp}: \mathcal{S}^{(1)} \times \cdots \times \mathcal{S}^{(i - 1)} \times \calD \rightarrow \mathcal{S}^{(i )}$ for $i \in [n]$. \\
	    \STATE Sample $I \xleftarrow{u.a.r.} [n]$
	    \STATE Let $\sigma_I(D) \leftarrow (d_I, d_2, \ldots, d_{I-1}, d_1, d_{I+1}, \ldots, d_n)$
       \FOR{$i \in [c]$}
       \STATE $s_i \leftarrow \calA^{(i)}_{ldp}(s_{1:i-1}; \sigma_I(D)[i])$
       \ENDFOR
       \STATE \textbf{return} sequence $s_{1:n}$
	\end{algorithmic}
	\label{alg:a_swap}
\end{algorithm}

\begin{thm}(Amplification by swapping)
For a domain $\calD$, let $\calA^{(i)}_{ldp}: \mathcal{S}^{(1)} \times \cdots \times \mathcal{S}^{(i - 1)} \times \calD \rightarrow \mathcal{S}^{(i )}$ for $i \in [n]$ (where $\mathcal{S}^{(i)}$ is the range space of $\calA^{(i)}_{ldp}$) be a sequence of algorithms s.t.
$\calA^{(i)}_{ldp}$ is $\eps_0$-DP for all values of auxiliary inputs in $\mathcal{S}^{(1)} \times \cdots \times \mathcal{S}^{(i - 1)}$. 
Let $\calA_{swap}: \calD^{n} \rightarrow \mathcal{S}^{(1)} \times \cdots \times \mathcal{S}^{(n)}$ be the algorithm that given a dataset $D = d_{1:n} \in \calD^{n}$, swaps the first element in $D$ with an element sampled u.a.r. in $D$, and then applies the local randomizers to the resulting dataset sequentially (see Algorithm~\ref{alg:a_swap}). 
 $\calA_{swap}$ satisfies $(\eps, \delta)$-DP at index 1 in the central model, for $\eps = \frac{e^{3\eps_0} (e^{\eps_0} - 1)^2}{2n} + e^{3\eps_0/2}(e^{\eps_0} - 1)\sqrt{\frac{2\log{(1/\delta)}}{n}}$.
Furthermore, if the $\calA^{(i)}$ are $(\eps_0,\delta_0)$-DP with $\delta_0 \leq \frac{(1 - e^{-\eps_0}) \delta_1}{4 e^{\eps_0} \left(2 + \frac{\ln(2/\delta_1)}{\ln(1/(1-e^{-5\eps_0}))}\right)}$, then $\calA_{swap}$ is $(\eps', \delta')$-DP with $\eps' = \frac{e^{24\eps_0} (e^{8\eps_0} - 1)^2}{2n} + e^{12\eps_0}(e^{8\eps_0} - 1)\sqrt{\frac{2\log{(1/\delta)}}{n}}$ and $\delta' = \delta + m (e^{\eps'}+1) \delta_1$.
\label{thm:ampl_swap}
\end{thm}

\begin{proof}
We start by proving the privacy guarantee of $\calA_{swap}$ for the case where for each $i \in [c]$, the local randomizer $\calA^{(i)}_{ldp}$ is $\eps_0$-DP, i.e., for the case where $\delta_0 = 0$. Let us denote the output sequence of $\calA_{swap}$ by $Z_1, Z_2, \ldots , Z_n$. Note that $Z_{1:n}$ can be seen as the output of a sequence of $n$ algorithms with conditionally independent randomness: $\calB^{(i)}: \mathcal{S}^{(1)} \times \cdots \times \mathcal{S}^{(i - 1)} \times \calD^{n} \rightarrow \mathcal{S}^{(i )}$ for $i \in [n]$. On input $s_{1:i-1}$ and $D$, $\calB^{(i)}$ outputs a random sample from the distribution of $Z_i|Z_{1:i-1}=s_{1:i-1}$. The outputs of $\calB^{(1)}, \ldots, \calB^{(i-1)}$ are given as input to $\calB^{(i)}$. Therefore, in order to upper bound the privacy parameters of $\calA_{swap}$, we analyze the privacy parameters of $\calB^{(1)}, \ldots, \calB^{(n)}$ and  apply the heterogeneous advanced composition for DP~\cite{KOV17}.

Next, observe that conditioned on the value of $I$, $Z_i$ is the output of $\calA^{(i)}_{ldp}(s_{1:i-1};d)$ with its internal randomness independent of $Z_{1:i-1}$. In particular, for $i \geq 2$, one can implement $\calB^{(i)}$ as follows. First, sample an index $T$ from the distribution of $I|Z_{1:i-1}=s_{1:i-1}$. 
Output $\calA^{(i)}_{ldp}(s_{1:i-1};d_1)$ if $T=i$, otherwise output $\calA^{(i)}_{ldp}(s_{1:i-1};d_i)$. For $\calB^{(1)}$, we first sample $T$ u.a.r. from $[n]$, and then output $\calA^{(1)}_{ldp}(d_T)$.

We now prove that for each $i\in[c]$, $\calB^{(i)}$ is $\left(\log\left(1 + \frac{e^{2\eps_0} (e^{\eps_0}-1)}{e^{2\eps_0} + (i - 1)  + (n - i)e^{\eps_0} } \right),0\right)$-DP at index 1. 
Let $D=d_{1:n}$ and $D'=(d'_1, d_{2:n})$ be 2 datasets differing in the first element. Let $s_{1:i-1}$ denote the input to $\calB^{(i)}$. Let $\mu$ be the probability distribution of $\calB^{(i)}(s_{1:i-1};D)$, and let $\mu_0$ (resp.\ $\mu_1$) be the distribution of $\calB^{(i)}(s_{1:i-1};D)$ conditioned on $T \neq i$ (resp.\ $T = i$). Let $q_i$ be the probability that $T=i$ (sampled from $I|Z_{1:i-1}=s_{1:i-1}$). By definition, $\mu = (1-q_i)\mu_0 + q_i \mu_1$.  Also, denote by $\mu', \mu_0'$, $\mu_1'$, and $q_i'$ the corresponding quantities when $\calB^{(i)}$ is run on $D'$. Thus, we get $\mu' = (1-q_i')\mu_0' + q_i' \mu_1'$.

For $i\in[n]$, we observe that $\mu_0=\mu_0'$, since in both cases the output is generated by $\calA^{(i)}_{ldp}(d_T)$ conditioned on $T \neq 1$ for $i = 1$, and $\calA^{(i)}_{ldp}(s_{1:i-1};d_i)$ for $i \geq 2$.
W.l.o.g. assume that $q_i \geq q_i'$.
Thus, we can shift $q_i - q_i'$ mass from the first component of the mixture in $\mu'$ to the second component to obtain
$$\mu' = (1-q_i) \mu_0 + q_i\left(\frac{q_i'}{q_i} \mu_1' + \left(1 - \frac{q_i'}{q_i}\right)\mu_0\right)
= (1-q_i) \mu_0 + q_i \mu_1''$$
This shows that $\mu$ and $\mu'$ are overlapping mixtures \cite{BBG18}.
Now, $\eps_0$-LDP of $\calA^{(i)}_{ldp}$ implies $\mu_0 \approxeq_{(\eps_0,0)} \mu_1$ and $\mu_0 \approxeq_{(\eps_0,0)} \mu_1'$. Moreover, $\eps_0$-LDP of $\calA^{(i)}_{ldp}$ also implies $\mu_1 \approxeq_{(\eps_0,0)} \mu_1'$, so by the joint convexity of the relation $\approxeq_{(\eps_0,0)}$ we also have $\mu_1 \approxeq_{(\eps_0,0)} \mu_1''$.
Thus, we can apply Advanced Joint Convexity of overlapping mixtures (Theorem 2 in \cite{BBG18}) to get that 
\begin{equation}
    \mu \approxeq_{(\log\left(1 + q_i(e^{\eps_0} - 1)\right),0)} \mu' \label{eqn:mus}
\end{equation}

We now claim that $q_i \leq \frac{e^{2\eps_0}}{e^{2\eps_0} + (i - 1)  + (n - i)e^{\eps_0} }$. Observe that for each $D^* \in \{ D, D'\}$, conditioning on $T=i$ reduces $\calA_{swap}$ to running $\calA^{(k)}_{ldp}, k \in [n]$ on $\sigma_i(D^*)$. Note that $\sigma_i(D^*)[1:i-1]$ differs from $\sigma_j(D^*)[1:i-1]$ in at most 2 positions for $j < i$, and at most 1 position for $j > i$. 
By $\eps_0$-LDP of $\calA^{(k)}_{ldp}, k \in [n]$, we get that
{\small
\begin{equation}
    \frac{\Pr[Z_{1:i-1}=s_{1:i-1} | T=i]}{\Pr[Z_{1:i-1}=s_{1:i-1} | T=j]} \leq e^{2\eps_0} \text{ for } j < i \quad \text{ and } \quad \frac{\Pr[Z_{1:i-1}=s_{1:i-1} | T=i]}{\Pr[Z_{1:i-1}=s_{1:i-1} | T=j]} \leq e^{\eps_0}  \text{ for } j > i
    \label{eqn:eps_01}
\end{equation} 
}
Now, on the lines of the proof of Lemma~\ref{lem:pr}, we have:
\begin{align*}
  & \frac{\Pr[Z_{1:i-1}=s_{1:i-1} | T=i]}{\Pr[Z_{1:i-1}=s_{1:i-1}]} \\
  & \qquad =   \frac{\Pr[Z_{1:i-1}=s_{1:i-1} | t=i]}{\sum\limits_{j = 1}^n\Pr[Z_{1:i-1}=s_{1:i-1}| T=j]\Pr[T=j]} \\
  & \qquad = \frac{1}{\sum\limits_{j = 1}^n\frac{\Pr[Z_{1:i-1}=s_{1:i-1}| t=j]}{\Pr[Z_{1:i-1}=s_{1:i-1} | t=i]}\Pr[T=j]} \\
  & \qquad = \frac{n}{1 + (i - 1) \sum\limits_{j < i}\frac{\Pr[Z_{1:i-1}=s_{1:i-1}| T=j]}{\Pr[Z_{1:i-1}=s_{1:i-1} | T=i]} + (n - i)\sum\limits_{k > i}\frac{\Pr[Z_{1:i-1}=s_{1:i-1}| T=k]}{\Pr[Z_{1:i-1}=s_{1:i-1} | T=i]} } \\
  & \qquad \leq \frac{n}{1 + (i - 1)e^{-2\eps_0}  + (n - i)e^{-\eps_0} } = \frac{ne^{2\eps_0}}{e^{2\eps_0} + (i - 1)  + (n - i)e^{\eps_0} }
\end{align*}
where the third equality follows as for every $j \in [n], \Pr[T=j] = \frac{1}{n}$, and the first inequality follows from inequality~\ref{eqn:eps_01}.

This immediately implies our claim, since 
\begin{align*}
    q_i & = \Pr[T=i | Z_{1:i-1}=s_{1:i-1}] = \frac{\Pr[Z_{1:i-1}=s_{1:i-1} | T=i] \cdot \Pr[T=i]}{\Pr[Z_{1:i-1}=s_{1:i-1}]} \\
    & \leq \frac{e^{2\eps_0}}{e^{2\eps_0} + (i - 1)  + (n - i)e^{\eps_0} }
\end{align*}
where the inequality follows from \eqref{eqn:zi}, and as $\Pr[T=i] = \frac{1}{n}$.
Substituting the value of $q_i$ in \eqref{eqn:mus}, we get that for each $i\in[n]$, algorithm $\calB^{(i)}$ is $\left(\eps_i,0\right)$-DP at index 1, where $\eps_i = \log\left(1 + \frac{e^{2\eps_0} (e^{\eps_0}-1)}{e^{2\eps_0} + (i - 1)  + (n - i)e^{\eps_0} }\right) $.  
This results in $\eps_i \leq \log\left(1 + \frac{e^{\eps_0}(e^{\eps_0}-1)}{n - (i-1) \left( 1 -  \frac{1}{e^{\eps_0}}\right)}\right)$, and using Lemma~\ref{lem:het} for the sequence of mechanisms  $\calB^{(1)}, \ldots, \calB^{(n)}$ by setting $a = e^{\eps_0}(e^{\eps_0}-1)$, $b= 1- \frac{1}{e^{\eps_0}}$, and $k = n$, we get that
algorithm $\calA_{swap}$ satisfies $\left(\eps, \delta\right)$-DP at index 1, for 
$\eps = \frac{e^{3\eps_0} (e^{\eps_0} - 1)^2}{2n} + e^{3\eps_0/2}(e^{\eps_0} - 1)\sqrt{\frac{2\log{(1/\delta)}}{n}}$.

The case $\delta_0 > 0$ uses the same argument based on Lemma~\ref{lem:cheu} used in the proof of Theorem~\ref{thm:ampl_rep_new}. This arguments allows us to reduce the analysis to the case $\delta_0 = 0$ and modify the final $\eps$ and $\delta$ accordingly.

\end{proof}

Now, we are ready to prove Theorem~\ref{thm:ampl_shuff}.

\begin{proof}[Proof of Theorem~\ref{thm:ampl_shuff}]
This proof proceeds in a similar manner as the proof of Theorem 7 in \cite{soda-shuffling}. Let $D$ and $D'$ be 2 datasets of length $n$ that differ at some index $i^* \in [n]$. Algorithm $\calA_{sl}$ can be alternatively seen as follows. Pick a random one-to-one mapping $\pi^*$ from $\{2, \ldots, n\} \rightarrow [n] \setminus \{i^*\}$ and let $\pi^*(D) = (d_{i^*}, d_{\pi^*(2)}, \ldots, d_{\pi^*(n)})$. Next, apply $\calA_{swap}$ to $\pi^*(D)$. It is easy to see that for a u.a.r. chosen $\pi^*$ and u.a.r. $I \in [n]$, the distribution of $\sigma_I(\pi^*(D))$ is a uniformly random permutation of elements in $D$.

For a fixed $\pi^*$, we know that $\pi^*(D)$ and $\pi^*(D')$ differ only in the element with index 1. Therefore, in the case $\delta_0 = 0$, from Theorem~\ref{thm:ampl_swap}, we get $\calA_{swap}(\pi^*(D)) \approxeq_{\eps, \delta} \calA_{swap}(\pi^*(D'))$ at index 1, for
 $\eps = \frac{e^{3\eps_0} (e^{\eps_0} - 1)^2}{2n} + e^{3\eps_0/2}(e^{\eps_0} - 1)\sqrt{\frac{2\log{(1/\delta)}}{n}}$,
 which implies $\calA_{sl}(D) \approxeq_{\eps, \delta} \calA_{sl}(D')$.

The case $\delta_0 > 0$ follows similarly from the corresponding setting of Theorem~\ref{thm:ampl_swap}.
\end{proof}

%% file: protocol_avg.tex
\subsection{Proof of Theorem~\ref{thm:ampl_avg}}

Let $L = (L_1, \ldots, L_m)$ represent the number of users contributing to each of the update steps, i.e., $L_i = |S_i|$ for $i \in [m]$.
We start by considering the output distribution of $\calA_{avg}(D)$ conditioned on $L = \ell$ for some $\ell \in [n]^m$ s.t. $\sum_i \ell_i = n$.
This distribution is the same as the one produced by Algorithm~\ref{alg:a_bins} with bin sizes $\ell$ on a random permutation $\pi(D)$ of the original dataset $D$.
To analyze the privacy of $\calA_{bin}(\pi(D), \ell)$ we use the reduction from shuffling to swapping \cite{soda-shuffling} %
.
This reduction says it suffices to analyze the privacy of $D \mapsto \calA_{bin}(\sigma(D), \ell)$ on a pair of datasets $D$ and $D'$ differing in the first record, where $\sigma(D)$ randomly swaps $d_1$ with $d_I$ for $I$ uniformly sampled from $[n]$.

\begin{algorithm}[ht]
\caption{$\calA_{bin}$: DP-SGD with Bins}
\begin{algorithmic}[1]
    \REQUIRE Dataset $D=d_{1:n}$, bin sizes $\ell \in [n]^m$ with $\sum_i \ell_i = n$, local randomizer $\calA_{ldp}$
    \STATE Initialize model $\theta_1 \in \mathbb{R}^p$
    \STATE $j \gets 1$
    \FOR{$i \in [m]$}
        \IF{$\ell_i = 0$}
            \STATE $\theta_{i+1} \gets \theta_{i}$
        \ELSE
            \STATE $\tilde{g}_i \gets 0$
            \FOR{$k \in \{j, \ldots, j + \ell_i - 1\}$}
                \STATE $\tilde{g}_i \gets \tilde{g}_i + \calA_{ldp}(d_k, \theta_{i})$
            \ENDFOR
            \STATE $j \gets j + \ell_i$
            \STATE $\theta_{i+1} \leftarrow \text{ModelUpdate}(\theta_{i}; \tilde{g}_i / \ell_i)$
        \ENDIF
    \ENDFOR
    \STATE \textbf{return} sequence $\theta_{2:m+1}$
\end{algorithmic}
\label{alg:a_bins}
\end{algorithm}

\begin{thm}\label{thm:binSGD}
Suppose $\calA_{ldp}: \calD \times \Theta \rightarrow \Theta$ is an $\eps_0$-DP local randomizer.
Let $\ell \in [m]^n$ with $\sum_i \ell_i = n$.
Also, for any dataset $D = \{d_1, \ldots, d_n\}$, define $\sigma(D)$ be the operation that randomly swaps $d_1$ with $d_I$ for $I$ uniformly sampled from $[n]$. 
For any $\delta \in (0,1)$, the mechanism $M(D) = \calA_{bin}(\sigma(D), \ell)$ is $(\eps,\delta)$-DP at index $1$ with $\eps = \frac{\ltwo{\ell}^2 e^{4 \eps_0} (e^{\eps_0}-1)^2}{2 n^2} + \frac{\ltwo{\ell} e^{2 \eps_0} (e^{\eps_0}-1)}{n} \sqrt{2 \log(1/\delta)}$.
Furthermore, if $\calA_{ldp}$ is $(\eps_0,\delta_0)$-DP with $\delta_0 \leq \frac{(1 - e^{-\eps_0}) \delta_1}{4 e^{\eps_0} \left(2 + \frac{\ln(2/\delta_1)}{\ln(1/(1-e^{-5\eps_0}))}\right)}$, then $M$ is $(\eps',\delta')$-DP with $\eps' = \frac{\ltwo{\ell}^2 e^{32 \eps_0} (e^{8\eps_0}-1)^2}{2 n^2} + \frac{\ltwo{\ell} e^{16 \eps_0} (e^{8\eps_0}-1)}{n} \sqrt{2 \log(1/\delta)}$ and $\delta' = \delta + m (e^{\eps'}+1) \delta_1$.

\end{thm}
\begin{proof}
Let $\sigma(D) = (\tilde{d}_1, \ldots, \tilde{d}_n)$ denote the dataset after the swap operation.
Using the bin sizes $\ell$, we split this dataset into $m_0$ disjoint datasets $\tilde{D}_1, \ldots, \tilde{D}_{m_0}$ of sizes $|\tilde{D}_i| = \ell_i$ with $\tilde{D}_1 = (\tilde{d}_1, \ldots, \tilde{d}_{\ell_1})$, and so on.
Note that each of the outputs is obtained as $\theta_{i+1} \gets \calA^{(i)}(\theta_{i}; \tilde{D}_i)$ with
\begin{align*}
    \calA^{(i)}(\theta_{i}; \tilde{D}_i) = \text{ModelUpdate}\left(\theta_{i}; \frac{1}{\ell_i} \sum_{\tilde{d} \in \tilde{D}_i} \calA_{ldp}(\tilde{d}, \theta_{i})\right) \enspace
\end{align*}
By post-processing, each of the $\calA^{(i)}$ is $(\eps_0, \delta_0)$-DP.

The next step is to modify these mechanisms to reduce the analysis to a question about adaptive composition.
Thus, we introduce mechanisms $\calB^{(i)}$ for $i \in [m_0]$ that take as input the whole dataset $D$ and the outputs $\theta_{1:i} = (\theta_1, \ldots, \theta_{i})$ of the previous mechanisms.
Mechanism $\calB^{(i)}$ starts by splitting the dataset $D$ into $m_0$ disjoint datasets ${D}_1, \ldots, {D}_{m_0}$ of sizes $|{D}_i| = \ell_i$ as above.
Then, it returns $\calA^{(i)}(\theta_{i}; \bar{D}_i)$ for a dataset $\bar{D}_i$ of size $\ell_i$ constructed as follows: with probability $p_i = \Pr[d_1 \in \tilde{D}_i | \theta_{1:i}]$ it takes $\bar{D}_i$ to be the dataset obtained by replacing a random element from $D_i$ with $d_1$, and with probability $1 - p_i$ it takes $\bar{D}_i = D_i$.
Note this construction preserves the output distribution since for any $\theta$ we have
{\small
\begin{align*}
    \Pr[\calA^{(i)}(\theta_{i}; \tilde{D}_i)  = \theta| \theta_{1:i}]
    & =
    (1-p_i) \Pr[\calA^{(i)}(\theta_{i}; D_i) = \theta | \theta_{1:i}, d_1 \notin \tilde{D}_i]\\
    &\qquad + \frac{p_i}{\ell_i} \sum_{d \in D_i} \Pr[\calA^{(i)}(\theta_{i}; D_i \cup \{d_1\} \setminus \{d\}) = \theta | \theta_{1:i}, d_1 \in \tilde{D}_i] \\
    &
    =
    \Pr[\calB^{(i)}(\theta_{1:i}; D) = \theta]
    \enspace
\end{align*}
}
To bound the probabilities $p_i$ we write:
\begin{align*}
    p_i
    &=
    \Pr[d_1 \in \tilde{D}_i | \theta_{1:i}]
    \\
    &=
    \frac{\Pr[\theta_{1:i} | d_1 \in \tilde{D}_i] \Pr[d_1 \in \tilde{D_i}]}{\Pr[\theta_{1:i}]}
    \\
    &=
    \frac{\ell_i}{n} \frac{\Pr[\theta_{1:i} | d_1 \in \tilde{D}_i]}{\sum_{k \in [m_0]} \Pr[\theta_{1:i} | d_1 \in \tilde{D}_k] \Pr[d_1 \in \tilde{D}_k]}
    \\
    &=
    \frac{\ell_i}{\sum_{k \in [m_0]} \ell_k \frac{\Pr[\theta_{1:i} | d_1 \in \tilde{D}_k]}{\Pr[\theta_{1:i} | d_1 \in \tilde{D}_i]}}
    \enspace
\end{align*}
To proceed, we assume $\delta_0 = 0$. If that is not the case, then the same argument based on Lemma~\ref{lem:cheu} used in the proof of Theorem~\ref{thm:ampl_rep_new} allows us to reduce the analysis to the case $\delta_0 = 0$ and modify the final $\eps$ and $\delta$ accordingly.
When the local randomizers satisfy pure DP, we have
\begin{align*}
    \sum_{k \in [m_0]} \ell_k \frac{\Pr[\theta_{1:i} | d_1 \in \tilde{D}_k]}{\Pr[\theta_{1:i} | d_1 \in \tilde{D}_i]}
    &\geq
    \ell_i + e^{-2 \eps_0} \sum_{k < i} \ell_k + e^{-\eps_0} \sum_{k > i} \ell_k
    \\
    &\geq
    e^{-2 \eps_0} n
    \enspace
\end{align*}
Thus we obtain $p_i \leq e^{2 \eps_0} \ell_i / n$.
Now, the overlapping mixtures argument used in the proof of Theorem~\ref{thm:ampl_rep_new} (see \cite{BBG18})
shows that $\calB^{(i)}$ is $\eps_i$-DP with
$\eps_i \leq \log(1 + e^{2 \eps_0} (e^{\eps_0} - 1) \ell_i / n)$.
Furthermore, the heterogenous advanced composition theorem~\cite{KOV17} implies that the composition of $\calB^{(1)}, \ldots, \calB^{(m_0)}$ satisfies $(\eps,\delta)$-DP with
\begin{align*}
    \eps
    &=
    \sum\limits_{i\in[k]} \frac{(e^{\eps_i} - 1) \eps_i}{e^{\eps_i} + 1} + \sqrt{2 \log{\frac{1}{\delta}} \sum\limits_{i\in[k]}\eps_i^2}
    \\
    &\leq
    \frac{(e^{\eps_0}-1)^2 e^{4 \eps_0} \ltwo{\ell}^2}{2 n^2}
    +
    \sqrt{\frac{2 (e^{\eps_0}-1)^2 e^{4 \eps_0} \ltwo{\ell}^2}{n^2} \log{\frac{1}{\delta}}}
\end{align*}
\end{proof}

To conclude the proof of Theorem~\ref{thm:ampl_avg}, we provide a high probability bound for $\ltwo{L}$ for random $L$ representing the loads of $m$ bins when $n$ balls are thrown uniformly and independently.

\begin{lem}\label{lem:hp_ampli_coef}
Let $L = (L_1, \ldots, L_m)$ denote the number of users checked in into each of $m$ update slots in the protocol from Figure~\ref{alg:a_dist_avg}.
With probability at least $1 - \delta$, we have $$\ltwo{L} \leq \sqrt{n + \frac{n^2}{m}} + \sqrt{n \log(1/\delta)}.$$
\end{lem}
\begin{proof}
The proof is a standard application of McDiarmid's inequality.
First note that $\ltwo{L}$ is a function of $n$ i.i.d.\ random variables indicating the bin where each ball is allocated.
Since changing the assignment of one ball can only change $\ltwo{L}$ by $\sqrt{2}$, we have
\begin{align*}
    \ltwo{L} \leq \mathbb{E} \ltwo{L} + \sqrt{n \log(1/\delta)}
\end{align*}
with probability at least $1-\delta$.
Finally, we use Jensen's inequality to obtain
\begin{align*}
    \mathbb{E}\sqbracket{\ltwo{L}}
    &\leq
    \sqrt{\mathbb{E} \sqbracket{\ltwo{L}^2}}
    =
    \sqrt{\sum_{i \in [m]} \mathbb{E}\sqbracket{ L_i^2}}
    =
    \sqrt{m \mathbb{E}\sqbracket{ \mathrm{Bin}(n, 1/m)^2}}
    \\
    &=
    \sqrt{m \left(\frac{n}{m} \bracket{1 - \frac{1}{m}} + \frac{n^2}{m^2}\right)}
    \leq
    \sqrt{n + \frac{n^2}{m}}
\end{align*}
\end{proof}

The privacy claim in Theorem~\ref{thm:ampl_avg} follows from using Lemma~\ref{lem:hp_ampli_coef} to condition with probability at least $1 -  \delta_2$ to the case where $L$ is such that
\begin{align*}
    \frac{\ltwo{L}}{n} \leq \sqrt{\frac{1}{n} + \frac{1}{m}} + \sqrt{\frac{\log(1/\delta_2)}{n}} \enspace,
\end{align*}
and for each individual event $L = \ell$ satisfying this condition, applying the analysis from Theorem~\ref{thm:binSGD} after the reduction from shuffling to averaging (see, e.g., the proof of Theorem~\ref{thm:ampl_shuff} below).

%% file: main.bbl
\begin{thebibliography}{10}

\bibitem{DP-DL}
M.~Abadi, A.~Chu, I.~J. Goodfellow, H.~B. McMahan, I.~Mironov, K.~Talwar, and
  L.~Zhang.
\newblock Deep learning with differential privacy.
\newblock In {\em Proc. of the 2016 {ACM} {SIGSAC} Conf. on Computer and
  Communications Security ({CCS}'16)}, pages 308--318, 2016.

\bibitem{apple_report}
D.~P.~T. Apple.
\newblock {Learning with privacy at scale}, 2017.

\bibitem{augenstein2019generative}
S.~Augenstein, H.~B. McMahan, D.~Ramage, S.~Ramaswamy, P.~Kairouz, M.~Chen,
  R.~Mathews, et~al.
\newblock Generative models for effective ml on private, decentralized
  datasets.
\newblock {\em arXiv preprint arXiv:1911.06679}, 2019.

\bibitem{BC19}
V.~Balcer and A.~Cheu.
\newblock Separating local {\&} shuffled differential privacy via histograms.
\newblock {\em CoRR}, abs/1911.06879, 2019.

\bibitem{BBG18}
B.~Balle, G.~Barthe, and M.~Gaboardi.
\newblock Privacy amplification by subsampling: Tight analyses via couplings
  and divergences.
\newblock In S.~Bengio, H.~M. Wallach, H.~Larochelle, K.~Grauman,
  N.~Cesa{-}Bianchi, and R.~Garnett, editors, {\em Advances in Neural
  Information Processing Systems 31: Annual Conference on Neural Information
  Processing Systems 2018, NeurIPS 2018, 3-8 December 2018, Montr{\'{e}}al,
  Canada}, pages 6280--6290, 2018.

\bibitem{privacy-blanket}
B.~Balle, J.~Bell, A.~Gascon, and K.~Nissim.
\newblock The privacy blanket of the shuffle model.
\newblock In {\em Advances in Cryptology---CRYPTO}, 2019.

\bibitem{BassilyFTT19}
R.~Bassily, V.~Feldman, K.~Talwar, and A.~G. Thakurta.
\newblock Private stochastic convex optimization with optimal rates.
\newblock In H.~M. Wallach, H.~Larochelle, A.~Beygelzimer,
  F.~d'Alch{\'{e}}{-}Buc, E.~B. Fox, and R.~Garnett, editors, {\em Advances in
  Neural Information Processing Systems 32: Annual Conference on Neural
  Information Processing Systems 2019, NeurIPS 2019, 8-14 December 2019,
  Vancouver, BC, Canada}, pages 11279--11288, 2019.

\bibitem{BST14}
R.~Bassily, A.~Smith, and A.~Thakurta.
\newblock Private empirical risk minimization: Efficient algorithms and tight
  error bounds.
\newblock In {\em Proc. of the 2014 IEEE 55th Annual Symp. on Foundations of
  Computer Science (FOCS)}, pages 464--473, 2014.

\bibitem{esa}
A.~Bittau, {\'{U}}.~Erlingsson, P.~Maniatis, I.~Mironov, A.~Raghunathan,
  D.~Lie, M.~Rudominer, U.~Kode, J.~Tinn{\'{e}}s, and B.~Seefeld.
\newblock Prochlo: Strong privacy for analytics in the crowd.
\newblock In {\em Proceedings of the 26th Symposium on Operating Systems
  Principles, Shanghai, China, October 28-31, 2017}, pages 441--459. {ACM},
  2017.

\bibitem{bonawitz2019towards}
K.~Bonawitz, H.~Eichner, W.~Grieskamp, D.~Huba, A.~Ingerman, V.~Ivanov,
  C.~Kiddon, J.~Konecny, S.~Mazzocchi, H.~B. McMahan, et~al.
\newblock Towards federated learning at scale: System design.
\newblock {\em arXiv preprint arXiv:1902.01046}, 2019.

\bibitem{bubeck2015convex}
S.~Bubeck.
\newblock Convex optimization: Algorithms and complexity.
\newblock {\em Foundations and Trends{\textregistered} in Machine Learning},
  8(3-4):231--357, 2015.

\bibitem{dpmixnets}
A.~Cheu, A.~Smith, J.~Ullman, D.~Zeber, and M.~Zhilyaev.
\newblock Distributed differential privacy via mixnets.
\newblock {\em CoRR}, abs/1808.01394, 2018.

\bibitem{mic}
B.~Ding, J.~Kulkarni, and S.~Yekhanin.
\newblock Collecting telemetry data privately.
\newblock In {\em Advances in Neural Information Processing Systems 30: Annual
  Conference on Neural Information Processing Systems 2017, 4-9 December 2017,
  Long Beach, CA, {USA}}, pages 3571--3580, 2017.

\bibitem{DJW13}
J.~C. Duchi, M.~I. Jordan, and M.~J. Wainwright.
\newblock Local privacy and statistical minimax rates.
\newblock In {\em 54th Annual {IEEE} Symposium on Foundations of Computer
  Science, {FOCS} 2013, 26-29 October, 2013, Berkeley, CA, {USA}}, pages
  429--438. {IEEE} Computer Society, 2013.

\bibitem{ODO}
C.~Dwork, K.~Kenthapadi, F.~McSherry, I.~Mironov, and M.~Naor.
\newblock Our data, ourselves: Privacy via distributed noise generation.
\newblock In {\em Advances in Cryptology---EUROCRYPT}, pages 486--503, 2006.

\bibitem{DMNS}
C.~Dwork, F.~McSherry, K.~Nissim, and A.~Smith.
\newblock Calibrating noise to sensitivity in private data analysis.
\newblock In {\em Proc. of the Third Conf. on Theory of Cryptography (TCC)},
  pages 265--284, 2006.

\bibitem{dwork2014algorithmic}
C.~Dwork and A.~Roth.
\newblock The algorithmic foundations of differential privacy.
\newblock {\em Foundations and Trends in Theoretical Computer Science},
  9(3--4):211--407, 2014.

\bibitem{esa++}
{\'{U}}.~Erlingsson, V.~Feldman, I.~Mironov, A.~Raghunathan, S.~Song,
  K.~Talwar, and A.~Thakurta.
\newblock Encode, shuffle, analyze privacy revisited: Formalizations and
  empirical evaluation.
\newblock {\em CoRR}, abs/2001.03618, 2020.

\bibitem{soda-shuffling}
{\'U}.~Erlingsson, V.~Feldman, I.~Mironov, A.~Raghunathan, K.~Talwar, and
  A.~Thakurta.
\newblock Amplification by shuffling: From local to central differential
  privacy via anonymity.
\newblock In {\em Proceedings of the Thirtieth Annual ACM-SIAM Symposium on
  Discrete Algorithms}, pages 2468--2479. SIAM, 2019.

\bibitem{rappor}
{\'U}.~Erlingsson, V.~Pihur, and A.~Korolova.
\newblock {RAPPOR}: Randomized aggregatable privacy-preserving ordinal
  response.
\newblock In {\em Proc. of the 2014 {ACM} Conf. on Computer and Communications
  Security ({CCS}'14)}, pages 1054--1067. ACM, 2014.

\bibitem{FMTT18}
V.~Feldman, I.~Mironov, K.~Talwar, and A.~Thakurta.
\newblock Privacy amplification by iteration.
\newblock In {\em 59th Annual IEEE Symp. on Foundations of Computer Science
  (FOCS)}, pages 521--532, 2018.

\bibitem{BPV20}
B.~Ghazi, R.~Pagh, and A.~Velingker.
\newblock Scalable and differentially private distributed aggregation in the
  shuffled model.
\newblock {\em CoRR}, abs/1906.08320, 2019.

\bibitem{FLO}
P.~Kairouz, H.~B. McMahan, B.~Avent, A.~Bellet, M.~Bennis, A.~N. Bhagoji,
  K.~Bonawitz, Z.~Charles, G.~Cormode, R.~Cummings, R.~G.~L. D'Oliveira, S.~E.
  Rouayheb, D.~Evans, J.~Gardner, Z.~Garrett, A.~Gasc{\'{o}}n, B.~Ghazi, P.~B.
  Gibbons, M.~Gruteser, Z.~Harchaoui, C.~He, L.~He, Z.~Huo, B.~Hutchinson,
  J.~Hsu, M.~Jaggi, T.~Javidi, G.~Joshi, M.~Khodak, J.~Konecn{\'{y}},
  A.~Korolova, F.~Koushanfar, S.~Koyejo, T.~Lepoint, Y.~Liu, P.~Mittal,
  M.~Mohri, R.~Nock, A.~{\"{O}}zg{\"{u}}r, R.~Pagh, M.~Raykova, H.~Qi,
  D.~Ramage, R.~Raskar, D.~Song, W.~Song, S.~U. Stich, Z.~Sun, A.~T. Suresh,
  F.~Tram{\`{e}}r, P.~Vepakomma, J.~Wang, L.~Xiong, Z.~Xu, Q.~Yang, F.~X. Yu,
  H.~Yu, and S.~Zhao.
\newblock Advances and open problems in federated learning.
\newblock {\em CoRR}, abs/1912.04977, 2019.

\bibitem{KOV17}
P.~Kairouz, S.~Oh, and P.~Viswanath.
\newblock The composition theorem for differential privacy.
\newblock {\em {IEEE} Trans. Inf. Theory}, 63(6):4037--4049, 2017.

\bibitem{KLNRS}
S.~P. Kasiviswanathan, H.~K. Lee, K.~Nissim, S.~Raskhodnikova, and A.~D. Smith.
\newblock What can we learn privately?
\newblock In {\em 49th Annual {IEEE} Symp. on Foundations of Computer Science
  (FOCS)}, pages 531--540, 2008.

\bibitem{census2}
Y.~Kuo, C.~Chiu, D.~Kifer, M.~Hay, and A.~Machanavajjhala.
\newblock Differentially private hierarchical count-of-counts histograms.
\newblock {\em {PVLDB}}, 11(11):1509--1521, 2018.

\bibitem{FL1}
B.~McMahan, E.~Moore, D.~Ramage, S.~Hampson, and B.~A. y~Arcas.
\newblock Communication-efficient learning of deep networks from decentralized
  data.
\newblock In {\em Proceedings of the 20th International Conference on
  Artificial Intelligence and Statistics, {AISTATS} 2017, 20-22 April 2017,
  Fort Lauderdale, FL, {USA}}, pages 1273--1282, 2017.

\bibitem{MRTZ18}
H.~B. McMahan, D.~Ramage, K.~Talwar, and L.~Zhang.
\newblock Learning differentially private language models without losing
  accuracy.
\newblock {\em CoRR}, abs/1710.06963, 2017.

\bibitem{MU}
M.~Mitzenmacher and E.~Upfal.
\newblock {\em Probability and Computing: Randomized Algorithms and
  Probabilistic Analysis}.
\newblock Cambridge University Press, 2005.

\bibitem{pichapati2019adaclip}
V.~Pichapati, A.~T. Suresh, F.~X. Yu, S.~J. Reddi, and S.~Kumar.
\newblock Adaclip: Adaptive clipping for private sgd.
\newblock {\em arXiv preprint arXiv:1908.07643}, 2019.

\bibitem{link}
R.~Rogers, S.~Subramaniam, S.~Peng, D.~Durfee, S.~Lee, S.~K. Kancha, S.~Sahay,
  and P.~Ahammad.
\newblock Linkedin's audience engagements api: A privacy preserving data
  analytics system at scale, 2020.

\bibitem{shamir2013stochastic}
O.~Shamir and T.~Zhang.
\newblock Stochastic gradient descent for non-smooth optimization: Convergence
  results and optimal averaging schemes.
\newblock In {\em International Conference on Machine Learning}, pages 71--79,
  2013.

\bibitem{smith2017interaction}
A.~Smith, A.~Thakurta, and J.~Upadhyay.
\newblock Is interaction necessary for distributed private learning?
\newblock In {\em 2017 IEEE Symposium on Security and Privacy (SP)}, pages
  58--77. IEEE, 2017.

\bibitem{song2013stochastic}
S.~Song, K.~Chaudhuri, and A.~D. Sarwate.
\newblock Stochastic gradient descent with differentially private updates.
\newblock In {\em 2013 IEEE Global Conference on Signal and Information
  Processing}, pages 245--248. IEEE, 2013.

\bibitem{TAB19}
O.~Thakkar, G.~Andrew, and H.~B. McMahan.
\newblock Differentially private learning with adaptive clipping.
\newblock {\em CoRR}, abs/1905.03871, 2019.

\bibitem{V85}
J.~S. Vitter.
\newblock Random sampling with a reservoir.
\newblock {\em ACM Trans. Math. Softw.}, 11(1):37–57, Mar. 1985.

\bibitem{WangBK19}
Y.~Wang, B.~Balle, and S.~P. Kasiviswanathan.
\newblock Subsampled renyi differential privacy and analytical moments
  accountant.
\newblock In {\em The 22nd International Conference on Artificial Intelligence
  and Statistics, {AISTATS} 2019, 16-18 April 2019, Naha, Okinawa, Japan},
  pages 1226--1235, 2019.

\bibitem{PFF}
Y.-X. Wang, S.~E. Fienberg, and A.~J. Smola.
\newblock Privacy for free: Posterior sampling and stochastic gradient monte
  carlo.
\newblock In {\em Proceedings of the 32nd International Conference on
  International Conference on Machine Learning - Volume 37}, ICML’15, page
  2493–2502. JMLR.org, 2015.

\bibitem{WLKCJN17}
X.~Wu, F.~Li, A.~Kumar, K.~Chaudhuri, S.~Jha, and J.~F. Naughton.
\newblock Bolt-on differential privacy for scalable stochastic gradient
  descent-based analytics.
\newblock In S.~Salihoglu, W.~Zhou, R.~Chirkova, J.~Yang, and D.~Suciu,
  editors, {\em Proceedings of the 2017 {ACM} International Conference on
  Management of Data, {SIGMOD}}, 2017.

\end{thebibliography}
